\documentclass{article}

\usepackage{microtype}
\usepackage{graphicx}
\usepackage{subfigure}
\usepackage{booktabs} 
\usepackage{multirow}
\usepackage{algpseudocode}

\usepackage{url}

\usepackage{breakurl}
\usepackage[breaklinks]{hyperref}

\usepackage[accepted]{arxiv_icml}

\usepackage{amsmath}
\usepackage{amssymb}
\usepackage{mathtools}
\usepackage{amsthm}

\usepackage[capitalize,noabbrev]{cleveref}

\theoremstyle{plain}

\theoremstyle{definition}

\theoremstyle{remark}

\usepackage[textsize=tiny]{todonotes}

\icmltitlerunning{What needs to go right for an induction head?}

\begin{document}

\twocolumn[
\icmltitle{What needs to go right for an induction head? \\ A mechanistic study of in-context learning circuits and their formation}

\icmlsetsymbol{equal}{*}

\begin{icmlauthorlist}
\icmlauthor{Aaditya K. Singh}{gatsby}
\icmlauthor{Ted Moskovitz}{gatsby}
\icmlauthor{Felix Hill}{gdm}
\icmlauthor{Stephanie C. Y. Chan}{equal,gdm}
\icmlauthor{Andrew M. Saxe}{equal,gatsby}
\end{icmlauthorlist}

\icmlaffiliation{gatsby}{Gatsby Computational Neuroscience Unit, University College London}
\icmlaffiliation{gdm}{Google DeepMind}

\icmlcorrespondingauthor{Aaditya K. Singh}{aaditya.singh.21@ucl.ac.uk}

\icmlkeywords{Mechanistic interpretability, transformers, causal manipulations, in-context learning, Machine Learning}



\vskip 0.3in
]




\printAffiliationsAndNotice{\icmlEqualContribution} 

\begin{abstract}
In-context learning is a powerful emergent ability in transformer models. Prior work in mechanistic interpretability has identified a circuit element that may be critical for in-context learning -- the induction head (IH), which performs a match-and-copy operation. During training of large transformers on natural language data, IHs emerge around the same time as a notable phase change in the loss. Despite the robust evidence for IHs and this interesting coincidence with the phase change, relatively little is known about the diversity and emergence dynamics of IHs. Why is there more than one IH, and how are they dependent on each other? Why do IHs appear all of a sudden, and what are the subcircuits that enable them to emerge? We answer these questions by studying IH emergence dynamics in a controlled setting by training on synthetic data. In doing so, we develop and share a novel optogenetics-inspired causal framework for modifying activations throughout training. Using this framework, we delineate the diverse and additive nature of IHs. By \textit{clamping} subsets of activations throughout training, we then identify three underlying subcircuits that interact to drive IH formation, yielding the phase change. Furthermore, these subcircuits shed light on data-dependent properties of formation, such as phase change timing, already showing the promise of this more in-depth understanding of subcircuits that need to ``go right'' for an induction head.
\end{abstract}

\section{Introduction}
\label{intro}

Large language models (LLMs) trained on internet scale corpora \cite{GPT3} showcase a remarkable ability to perform \textit{in-context learning} (ICL), adapting to new inputs and tasks at test time. Given the increasing prevalence of LLMs, safety researchers have sought to understand the mechanisms underlying important phenomena like ICL. One approach in the service of this goal is \textit{mechanistic interpretability}: reverse engineering the computations performed by a model. \citet{InductionHeads} identified the induction circuit, which may be responsible for many of transformers' ICL abilities. 

\begin{figure*}[ht]
    \centering
    \includegraphics[width=0.95\textwidth]{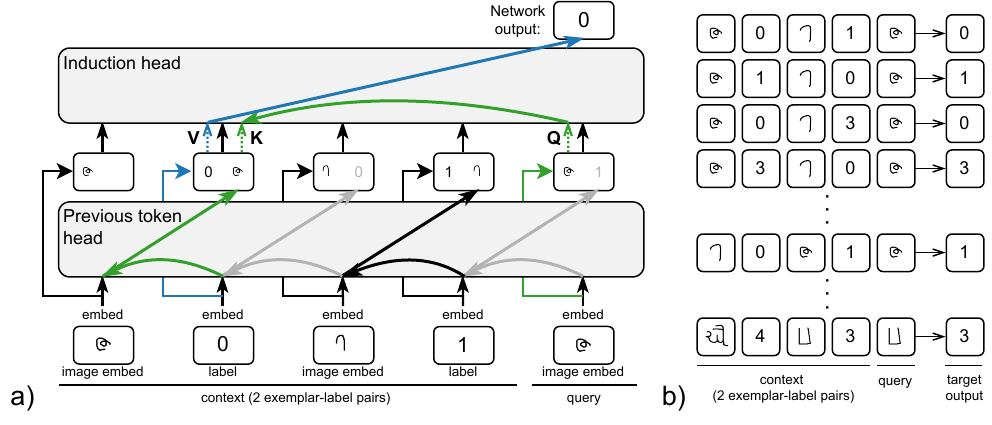}
    \vspace{-2em}
    \caption{\textbf{a)} Schematic of an induction circuit, involving a previous token head in Layer 1 and an induction head in Layer 2. The side-by-side labels and exemplars in the residual stream after Layer 1 are meant to indicate that information about both is superimposed (perhaps in different subspaces). We highlight\footnotemark the ``matching'' (green) and ``copying'' (blue) operations that span the two layers. Historically, focus has been devoted to the ``match'' operation. One of our key results is to demonstrate the important interactions from the ``copy'' operation. \textbf{b)} Example training sequences built from the Omniglot dataset and inspired by classical few-shot meta-training. The context consists of two exemplar-label pairs, where the exemplars are from different classes. The query exemplar comes from the same class as one of the exemplars in context. The in-context labels are randomly chosen. Every exemplar can appear with every possible label in every possible position, forcing the transformer to use ICL to minimize the training loss. Validation sequences either use held out class exemplars or held out pairs of labels.}
    \label{fig:ih_schematic_methods}
    \vspace{-1em}
\end{figure*}

An induction circuit is a two-layer circuit (Figure~\ref{fig:ih_schematic_methods}a) comprised of a ``previous token head'' in an earlier layer, followed by the eponymous ``induction head''. Previous token heads are attention heads responsible for attending to the previous token and copying it into the attending token's residual stream.\footnote{The residual stream was a term introduced by \citet{anthropicMathFramework} and refers to the embeddings after each layer that further layers ``read'' input from. Layers can be viewed as reading from and writing back into this stream. The term ``stream'' is meant to emphasize the residual skip connections.} Induction heads then perform a match-and-copy operation, looking for a match between a query derived from the current token and key derived from the output of the previous token head.
Critically, induction circuits typically emerge in a sharp phase change in the loss (e.g., Figure~\ref{fig:induction_heads}), which is often believed to be due to the interaction between two heads in different layers \cite{nanda_dynalist}. Presumably, neither the previous token nor the induction head are useful on their own for minimizing the loss.

Recent empirical work on ICL in transformers has introduced more nuances \cite{min2022rethinking, wei2023larger}, with \citet{singh2023transient} even finding that ICL can be a transient phenomenon that disappears with overtraining. These results establish a need to more thoroughly understand induction circuits, especially the \textit{dynamics} of formation during training. \citet{reddy2023mechanistic} begins to address this question using \textit{progress measures}, which track correlational relationships between intermediate activations during training. We instead use a causal approach, termed \textit{clamping}, which allows us to directly determine the circuits and dynamics that are causally affecting formation. Through these experiments, we bring to light a new set of mechanisms governing learning dynamics.

In this work, we take inspiration from the field of optogenetics in neuroscience, which allows precise causal manipulations of neural activity \cite{boyden_millisecond-timescale_2005}. Here, we develop a novel framework for causally modifying activations throughout model training. These causal-through-training clamping analyses extend beyond prior work, allowing us to easily study interactions between subcircuits and isolate the underlying factors that drive induction circuit formation. We utilize this framework to analyze induction circuit formation in transformers trained to solve a simple few-shot learning (FSL) task inspired by prior work \cite{Chan2022, omniglot}. As with any work towards a mechanistic understanding of dynamics, we first provide a thorough investigation of the induction circuits in a network trained on this task, shedding light on the dependence (and independence) of induction heads upon each other (Section~\ref{sec:ih_classification}). We find that multiple induction heads form, contributing additively to minimize the loss. Furthermore, we observe emergent redundancy, despite not applying regularization techniques such as dropout \cite{dropout}, mirroring findings on larger-scale language models \cite{sixteen_heads, voita-etal-2019-analyzing}. We also find the wiring between induction heads and previous token heads to be many-to-many, rather than one-to-one. Next, we focus on the dynamics of formation (Section~\ref{sec:subcircuits}), where we use clamping (causal manipulations of activations throughout training) to identify three, smoothly evolving underlying subcircuits whose interaction may be causing the seemingly discontinuous phase change.
Specifically, the third key evolving subcircuit is responsible for copying input labels to the output (highlighted in blue, Figure~\ref{fig:ih_schematic_methods}a), a function that was often believed to be easy (with most prior work focusing on the ``match'' operation instead, where an induction circuit has to find the right token to attend to). 
Finally, we show how data properties influence the timing of the phase change, and how this shift in induction circuit formation can be better understood by individually understanding the data-dependent formation of each of our identified subcircuits.

\footnotetext{We also gray out the previous token head operation for copying label tokens. This behavior does not robustly emerge as it isn't necessary since we only train on sequences of a fixed length.}

As a resource to the community, we open-source our codebase at \url{https://github.com/aadityasingh/icl-dynamics}, providing a tool for modifying activations throughout training and conducting further causal analyses of transformers' circuit elements. Our work and tooling represents an important step in mechanistic understanding of training dynamics, applied here to induction circuit formation, and we hope it spurs further progress on understanding how different computations in transformers are learned. 

\section{Methods}
\label{sec:methods}

\subsection{Experimental setup}
\label{sec:setup}

\begin{figure*}[ht]
    \centering
    \includegraphics[width=0.7\textwidth]{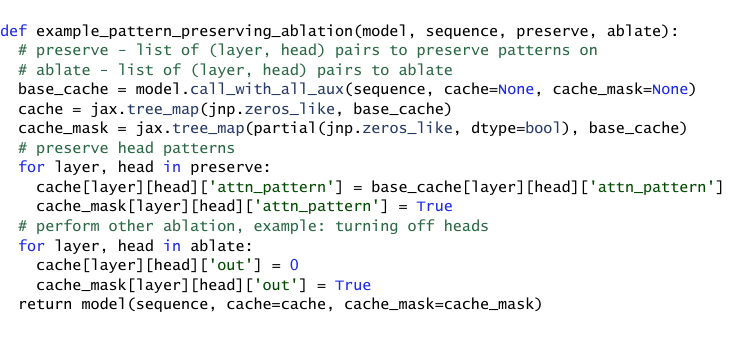}
    \vspace{-1.5em}
    \caption{Example pseudocode demonstrating a pattern preserving ablation using our framework.}
    \label{fig:example_framework}
    \vspace{-1em}
\end{figure*}

We train transformer models on a FSL task. Sequences are series of exemplar-label pairs, followed by a query exemplar (see Figure~\ref{fig:ih_schematic_methods}b). Exemplars come from the Omniglot dataset \cite{omniglot}, a common benchmark for FSL also used by prior work on ICL in transformers \cite{Chan2022, singh2023transient}. Omniglot consists of 1600 classes of 20 exemplars. Most of our experiments use a simplified dataset with a random set of $C=50$ classes and $E=1$ exemplar per class, which we found sufficient to elicit the relevant phenomena. To obtain input embeddings, we feed Omniglot images through an ImageNet-pretrained, frozen Resnet18 encoder \cite{resnet, imagenet}.

To isolate FSL capabilities, we use the standard meta-training setup where labels are randomized for each sequence. Specifically, each sequence can be viewed as a 2-way, 1-shot classification problem (see Figure~\ref{fig:ih_schematic_methods}b). Labels are randomly selected from $L$ one-hot labels. To obtain input embeddings, we use a standard learnt embedding layer.

In our default setting\footnote{ In Section~\ref{sec:data_dependent}, we vary these parameters to more deeply study the dependence of induction head formation on data properties.} ($C=50, E=1, L=5$), we have a total of 78400 unique training sequences (Appendix~\ref{appx:size_calculation}). On this data, we train causal, 2-layer attention-only transformer models, as used by prior work \cite{InductionHeads}, to study induction circuit formation. Model and optimizer hyperparameters can be found in Appendix~\ref{appx:model}.

To test generalization, we consider two test sets over held-out classes and relabelings. Specifically, for ``test (exemplars)'', we evaluate performance on a random, held-out set of $C_{test}=100$ classes, which verifies the generality of the ``match'' operation of induction circuits. For ``test (relabel)'', we hold out and test on a fixed percentage (20\%) of pairs of labels. This ensures that, while all labels are seen during training, not all pairs are. This test set verifies the generality of the ``copy'' operation of induction circuits.

\subsection{An artificial-optogenetics framework}
\label{sec:opto}

One of the contributions of this work is a novel training and analysis framework that easily exposes activations, and allows causal manipulations throughout training (as opposed to only after training, as in prior work). The framework enables a wide range of analyses---in this work, we use it to conduct targeted analyses of induction circuit formation via clamping. Our codebase is implemented in Equinox/JAX \cite{equinox, jax}, natively supports up to 50x speed-ups with \texttt{jax.jit},\footnote{See Appendix \ref{appx:cost} for details.} and is open-source.

In contrast to the standard practice of having modules with a forward function implemented as \texttt{\_\_call\_\_}, we have an underlying \texttt{call\_with\_all\_aux} method which returns a pytree of all intermediate activations. The \texttt{\_\_call\_\_} method wraps \texttt{call\_with\_all\_aux} and returns just the output. During the training process, \texttt{\_\_call\_\_} is used. For analysis, \texttt{call\_with\_all\_aux} can be easily used to expose all intermediate activations.

To allow for editing/ablating activations throughout training (or just post-hoc, as done in prior work), \texttt{call\_with\_all\_aux} accepts \texttt{cache} and \texttt{cache\_mask} arguments, which have the same shape as the output pytree of  \texttt{call\_with\_all\_aux}. These caches can be used to insert activations into the network, e.g. clamping activations at particular values during training. Combined with the functional, automatic differentiation of JAX, these caches can be input- and model-dependent, while still allowing for proper gradient routing. For example, Figure~\ref{fig:example_framework} shows how easy it is to implement the ``pattern-preserving\footnote{Here ``patterns'' are used to refer to the post-softmax attention scores from one token to others.} ablation'' of prior work \cite{InductionHeads}. Importantly, our framework allows implementing such causal manipulations \emph{throughout training}, which enables us to isolate the \emph{dynamics} of subcircuit formation (Section~\ref{sec:clamping}). While prior work\footnote{We note that some prior work in vision \cite{ranadive2023special} or masked language models \cite{chen2024sudden} has attempted to causally manipulate learning, but primarily through the use of regularizing losses. Our framework permits more direct interventions (on activations) throughout training.} was restricted to conducting mechanistic analyses on checkpoints from training, providing only correlational evidence, our framework allows \textit{causal} interventions on learning dynamics (see Appendix~\ref{appx:clamp_vs_progress} for further discussion).

\begin{figure}[t]
    \centering
    \includegraphics[width=\columnwidth]{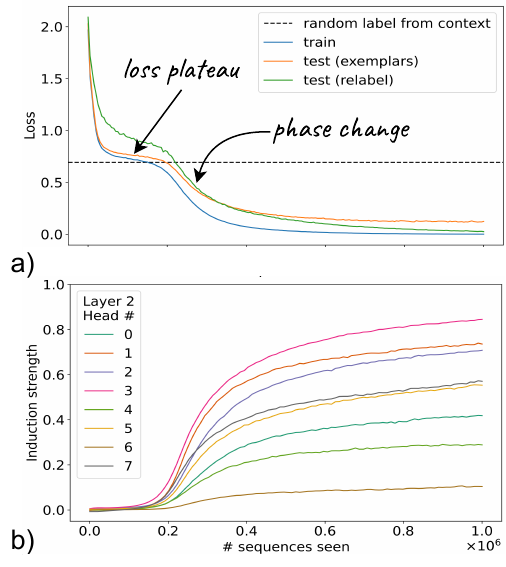}
    \vspace{-1.5em}
    \caption{\textbf{a)} Train and test loss curves. Transformers exhibit strong generalization to unseen classes (orange) and label pairs (green). The loss dynamics reveal a plateau (which may be indicative of a saddle point), where the model is randomly guessing between the two labels present in context (so it has 50\% accuracy, instead of the chance level of 20\% when there are $L=5$ labels). Then, there's a phase change in the loss which corresponds to the formation of induction circuits, reproducing the finding of \citet{InductionHeads}. \textbf{b)} Induction head strength for each Layer 2 head plotted over time. Induction head strength is defined as the attention weight given to the correct label token minus that to the incorrect token. All heads appear to have some induction-like behavior, with Head 3 being the strongest and emerging first.}
    \label{fig:induction_heads}
    \vspace{-2em}
\end{figure}

\section{Induction head emergence and diversity}
\label{sec:ih_classification}

We first establish that training transformers as described in Section \ref{sec:setup} leads to ICL abilities and induction head emergence. Figure~\ref{fig:induction_heads}a shows loss dynamics. Initially, the transformer reaches a plateau (perhaps indicative of a saddle point) where its accuracy is $\approx$50\% (a loss of $\log 2$). This corresponds to the network having learned to randomly select from the two labels in context,\footnote{The model tends to place equal weight on the two labels, rather than randomly putting high weight on one.} rather than the full set of all possible labels ($L=5$). Then, a sudden phase change in the loss leads to near-0 loss and near-perfect accuracy on the task (99.99\%). At this point, we also observe strong generalization to unseen exemplars and unseen label pairs, indicating the transformer has learnt a general ICL mechanism and is not simply memorizing the training sequences.

We find that the phase change in the loss corresponds to the formation of induction circuits. We measure the ``induction strength'' of a head, a common progress measure, as a difference in attention weights from the query token: [attention to the correct in-context label token] - [attention to the incorrect in-context label token]. For example, on the sequence ``A 0 B 1 A'', this would be the difference [attention weight from 5th token to 2nd token] - [attention weight from 5th token to 4th token]. We see that the emergence of induction heads corresponds to the phase change in the loss.

To verify that Layer 2 induction heads instead of Layer 1 heads are primarily contributing to task performance, we use our artificial optogenetics framework to ablate the residual connection between Layer 1 heads and the output. We do this by setting Layer 2 activations (attention patterns and values vectors) to those from an un-ablated forward pass (keeping induction circuits intact) and then ablating Layer 1 head outputs. 
Such an ablation leads to a negligible change in loss and accuracy (which drops by 0.01\% to 99.98\%), indicating that Layer 1 heads are not responsible for task performance. When we apply this ablation across checkpoints, we see that earlier in training, Layer 1 heads are contributing to minimizing the loss \citep[perhaps through the use of skip-trigrams;][]{anthropicMathFramework}, but no longer towards the end (Appendix Figure~\ref{fig:ablate_l1_checkpoints}).

\begin{figure}[t]
    \centering
    \includegraphics[width=\columnwidth]{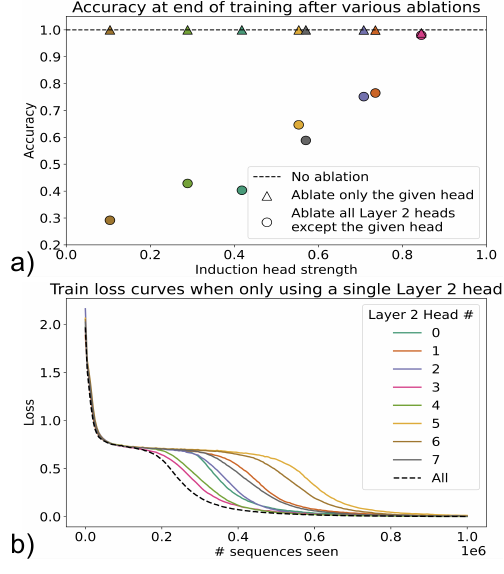}
    \vspace{-1.7em}
    \caption{\textbf{a)} Effect of various ablations on accuracy (effects on loss are shown in Appendix Figure \ref{fig:additive_heads_loss}). Ablating any single head (triangles) leads to virtually no decrease in task performance, with the exception of Head 3, which leads to a 1\% decrease. Ablating all but a specific head (circles) isolates how useful that specific head is, which correlates well to the induction strength (x-axis), the metric from Figure~\ref{fig:induction_heads}b. Importantly, ablating Head 3 (pink triangle) performs very similar to ablating everything except Head 3 (pink circle), which indicates the other heads function additively, and together can make up for the deletion of Head 3. \textbf{b)} Training loss curves when training from scratch with only a single head from Layer 2 active (and the rest ablated). Black dotted line is the loss profile from the training run in Figure~\ref{fig:induction_heads}. Colors chosen to match Figure~\ref{fig:induction_heads}b. Each Layer 2 head on its own can learn to solve the task, though the timing of the phase change shifts and learning is slower.}
    \label{fig:additive_heads}
    \vspace{-1.5em}
\end{figure}

\subsection{The additive nature of induction heads}
\label{sec:additive_heads}
One of the notable features of Figure~\ref{fig:induction_heads}b is the fact that many heads become strong induction heads. What purpose do all these heads serve? Are they necessary to solve the task?

To answer these questions, we consider two types of ablations of the fully trained network (Figure~\ref{fig:additive_heads}a). The first type of ablation is similar to prior mechanistic work \cite{InductionHeads}, where we ablate a given head and observe the effect on network performance (triangles, Figure~\ref{fig:additive_heads}a). However, we found that this type of ablation was not very informative, and even potentially misleading, as ablating any head except the strongest leads to an almost unobservable decrease in task performance. Next, we considered an ablation inspired by work on head pruning \cite{sixteen_heads}, where all Layer 2 heads are turned off and only one head is turned on (circles, Figure~\ref{fig:additive_heads}a). This ablation allows us to identify the positive contribution of each head to the 
task.
\footnote{There are some intuitive connections to the commonly used \textit{logit lens} \cite{logit_lens}. However, we emphasize that our method relies on causal ablations only and takes into account LayerNorm, which is often a thorn for interpretability research. We measure task performance with the ablation applied, instead of a change in logits.}
As seen in Figure~\ref{fig:additive_heads}a, the latter ablation (circles) reveals a clearer picture of the importance of induction heads, showing the correlation between induction head strength (which is calculated using attention scores), and task performance.

We see that despite Head 3 being able to mostly solve the task on its own (achieving an accuracy of 98\%), other single heads can still achieve strong performance. Furthermore, heads seem to have an additive effect, most clearly demonstrated\footnote{We provide further evidence in Appendix~\ref{appx:extra_figures} Figure~\ref{fig:additive_heads_one_at_a_time}.} by comparing the pink triangle (which corresponds to ablating Head 3) and pink circle (which corresponds to only keeping Head 3). Both achieve around 98\% accuracy, indicating that the other, weaker heads together can make up for the loss of Head 3. 
These results echo similar findings in head pruning \cite{sixteen_heads} and layer-wise redundancy of language models \cite{mcgrath2023hydra}.\footnote{Such redundancy is also a common observation in neuroscience \cite{hennig2018constraints}.}
Our analysis builds on this work and cautions against the use of purely knock-out ablations in mechanistic analyses, which may overlook redundancies and incorrectly dismiss network pieces as not implementing a certain function.

\subsection{Networks use additional capacity for faster training, even if it is not necessary}
\label{sec:networks_use_capacity}

Given the additive qualities observed above, a natural question is---does the network need multiple induction heads to \textit{learn} to solve the task? We know that we can ablate many heads and not hurt performance \textit{at the end of training}, but perhaps these heads play a crucial role \textit{during training}?
To answer these question, we use our artificial optogenetics framework to ablate all Layer 2 heads but one \textit{throughout training}. We do this for each Layer 2 head and show train loss curves in Figure~\ref{fig:additive_heads}b (all networks trained from the same random initialization).
We find that final networks attain the same near-perfect train performance as the network trained with 8 heads in Layer 2 (black dotted line in Figure~\ref{fig:additive_heads}b, blue line in Figure~\ref{fig:induction_heads}), but with slightly slower dynamics (delayed phase change timing). Thus, it seems the network may make use of the additional capacity during training to learn faster, even though it is not strictly necessary for the task.

These experiments connect to notions of lottery tickets\footnote{Lottery tickets typically refer to sub-networks sufficient to solve the task determined largely by initialization.} \cite{lottery_ticket, nandaDynamicsProblems}. Specifically, we note that different Layer 2 heads exhibit phase changes at different times (but all eventually learn to solve the task). Head 3 is the ``quickest to learn'' in Figure~\ref{fig:additive_heads}b, which may also be the reason it becomes the strongest when training with all heads (Figure~\ref{fig:induction_heads}b)---perhaps heads are racing to minimize the loss, similar to the mechanism in \citet{saxe2022neural}. Once Head 3 emerges, though, the ordering of phase changes in Figure~\ref{fig:additive_heads}b does not match emergence in the full network (Figure~\ref{fig:induction_heads}b)---e.g., Head 1 is the second to emerge and second strongest, but the third slowest when trained on its own. Additive interactions could be causing Head 1 to be learned sooner, due to the presence of Head 3 (we further explore such interactions in Appendix \ref{sec:residual_heads}).

\subsection{Previous token heads influence induction heads in a many-to-many fashion}
\label{sec:previous_token}
Prior work \cite{InductionHeads} has stressed the two-layer nature of induction circuits, where induction heads in later layers rely on the output of previous token heads in earlier layers to attend to the correct token. By inspecting attention patterns, we identify three previous token heads in Layer 1: Heads 1, 2, and 5. Through a series of ablation analyses, we find that each of these heads is enough to elicit above-chance accuracy in at least one of the strongest induction heads in Layer 2 (see Appendix Figure~\ref{fig:pt_indiv_contrib}b). These results indicate a many-to-many wiring between previous token heads and induction heads, with previous token heads operating redundantly (just as induction heads operate redundantly to solve the task). Full details are provided in Appendix \ref{appx:layer1}.

\section{Three interacting subcircuits give rise to the phase change in induction head formation}
\label{sec:subcircuits}

We now turn our attention to the dynamics of induction circuit formation. Prior work \cite{jermyn2022Scurves} indicates that reverse-S phase changes in the loss are often due to multiple interacting components. Below, we delineate the necessary computations that comprise an induction circuit, then study their formation dynamics in isolation using our artificial optogenetics framework.

\subsection{Terminology for this section}
\label{sec:ih_steps}

For clarity, we isolate and define the 5 key computations involved in an induction circuit, as illustrated in Figure~\ref{fig:subcircuits}c. In Section \ref{sec:clamping}, we'll show that these 5 computations can be grouped into the 3 primary interacting subcircuits.

\textbf{Step 1 (PT-attend)} 
\phantomsection
\label{step:1}
Layer 1 head attends to previous token.

\textbf{Step 2 (PT-copy)} 
\phantomsection
\label{step:2}
Layer 1 head copies previous token value into the current token's residual stream.

\textbf{Step 3 (Routing Q/K/V)}
\phantomsection
\label{step:3}
Layer 2 head reads Q/K/V\footnote{Q/K/V are the query, key, and value in the attention computation of a transformer \cite{vaswani2017attention}.} from the correct subspaces of the residual stream: Q from the residual, K from the output of the Layer 1 head, and V from the residual or from other Layer 1 heads.\footnote{Unlike \citet{InductionHeads}, we observe non-trivial V-composition in our networks. See Appendix \ref{appx:layer1} for details.} 

\textbf{Step 4 (IH-Match)}
\phantomsection
\label{step:4}
Layer 2 head matches Q to ``same'' K. 

\textbf{Step 5 (IH-copy)}
\phantomsection
\label{step:5}
Layer 2 copies the value to the output.

Steps 1 and 2 comprise what's canonically known as a previous token head, while Steps 3-5 form the induction head. The ``match operation'' (Figure~\ref{fig:ih_schematic_methods}a, blue) consists of Steps 1, 2, 3qk, 4. The ``copying operation'' (Figure~\ref{fig:ih_schematic_methods}a, green) consists of Steps 3v, 5.

\subsection{\textit{Clamping} computations to understand the causal effects of subcircuits on dynamics}
\label{sec:clamping}

Now that we've defined these computations, we ask: how do the learning dynamics of each of these computations causally influence the learning dynamics of the full model? Can we explain seemingly discontinuous qualitative phase changes, such as the induction circuit formation, in terms of subcircuits with smoother, exponential learning dynamics?

To motivate our analysis, we extend the toy model from \citet{jermyn2022Scurves} to the case of three interacting vectors.\footnote{Our tensor product formalism for analyzing sub-circuits could be extended beyond 3. We chose 3 as we ended up finding 3 primary subcircuits contributing to induction circuit formation. We also assume that $\mathbf{a}^{true},\mathbf{b}^{true},\mathbf{c}^{true} \neq \mathbf{0}$.} Specifically, we have three vectors $\mathbf{a}$, $\mathbf{b}$, and $\mathbf{c}$ that are learnt using a simple mean-squared error loss, corresponding to learning their values via a tensor product:
\begin{align*}
    \mathcal{L}(\mathbf{a}, \mathbf{b}, \mathbf{c}) = \frac{1}{2}\sum_{i,j,k}\left(a^{true}_i b^{true}_j c^{true}_k - a_i b_j c_k\right)^2
\end{align*}
The interaction between the evolution of $\mathbf{a}$, $\mathbf{b}$, and $\mathbf{c}$ gives rise to a phase change in dynamics (black, Figure~\ref{fig:subcircuits}a) following a loss plateau caused by a saddle point (see proof in Appendix~\ref{appx:saddle_point}). If we clamp $\mathbf{c}$ to its final value, learning is faster (blue, Figure~\ref{fig:subcircuits}a), but the co-evolution of $\mathbf{a}$ and $\mathbf{b}$ still results in a phase change. If we clamp $\mathbf{b}$ \textit{and} $\mathbf{c}$ (purple, Figure~\ref{fig:subcircuits}a), we isolate a smooth exponential loss curves corresponding to the formation dynamics of $\mathbf{a}$. 
This toy model gives us the intuition that phase changes may be caused by two or more subcircuits that, when learned on their own, evolve exponentially, but when co-evolving, induce phase changes. Furthermore, these subcircuits could be isolated by clamping all the other interacting components throughout training. Notably, it's not clear how to use existing interpretability techniques, such as progress measures \cite{nanda2023progress, reddy2023mechanistic}, to uncover these subcircuits (Appendix \ref{appx:clamp_vs_progress}).

\begin{figure}
    \centering
    \includegraphics[width=\columnwidth]{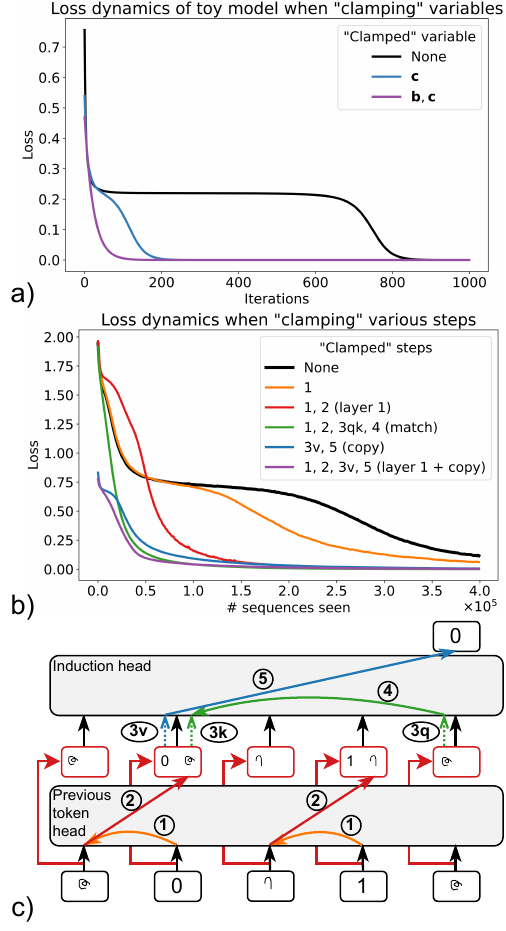}
    \vspace{-1.5em}
    \caption{\textbf{a)} Loss dynamics when clamping various variables in the toy model presented in Section~\ref{sec:clamping}. Black shows the learning dynamics when no variable is clamped. Only when all other interacting components ($\mathbf{b}, \mathbf{c}$) are clamped does the loss curve become exponential. \textbf{b)} Loss dynamics when clamping various computations outlined in Section~\ref{sec:ih_steps}. Black shows the training dynamics of the full network with nothing clamped. \textbf{c)} Induction circuit schematic (from Figure \ref{fig:ih_schematic_methods}a), with computation steps labeled. Arrow colors chosen to illustrate which steps are additionally clamped.}
    \label{fig:subcircuits}
    \vspace{-1.5em}
\end{figure}

We apply these insights from the toy model to understanding the key underlying subcircuits in an induction circuit. Specifically, we aim to isolate subcircuits where each individual subcircuit does not exhibit a phase change in its learning dynamics, like variables $\mathbf{a}, \mathbf{b},$ and $\mathbf{c}$ in the toy model. We restrict ourselves to considering the induction circuit with induction head Layer 2 Head 3, a minimal setting where we observe a phase change in the loss dynamics (see Figure~\ref{fig:additive_heads}b). In Section~\ref{sec:data_dependent}, we will show how understanding these subcircuits can help us understand the data-dependence of phase change timing, affirming the importance of discovering smoothly-learned underlying subcircuits.

To isolate these subcircuits, we iteratively clamp the computations delineated in Section~\ref{sec:ih_steps} at the activation level,\footnote{Specifics of our implementation are provided in Appendix \ref{appx:clamps_used} and why fixing weights to those at the end of training is not sufficient (see Appendix \ref{appx:clamp_routing_complicated}).} using our artificial optogenetics framework. Results are shown in Figure~\ref{fig:subcircuits}b. We start by clamping Step  \hyperref[step:1]{1}
(PT-attend), the previous token head attention pattern (an oft-used progress measure for IH formation). We train a network where one\footnote{We also conducted experiments where multiple previous token heads were provided and observed similar loss dynamics. See Appendix \ref{appx:clamps_used}.} of the Layer 1 heads is clamped from the start of training to a perfect PT-attend pattern. We see the loss dynamics still exhibit a phase change (orange), with a similar profile, indicating the rest of the computations (\hyperref[step:2]{2}-\hyperref[step:5]{5}) contain interacting subcircuits. Next, we clamp the full Layer 1 computation (PT-attend \textit{and} PT-copy) and ensure that the superimposed residual and output of Layer 1 are disentangled (see Appendix~\ref{appx:clamps_used} for details). We find that the dynamic profile of the loss shifts substantially (red). While a phase change is still observed, it is much sharper, lasting only $\approx7.5e4$ iterations (compared to $2e5$, black). This suggests that the formation of Layer 1 attention+output circuits is likely a key sub-component, but there are still interacting sub-components in Steps \hyperref[step:3]{3}, \hyperref[step:4]{4}, and \hyperref[step:5]{5}.

\begin{figure*}
    \centering
    \includegraphics[width=0.97\textwidth]{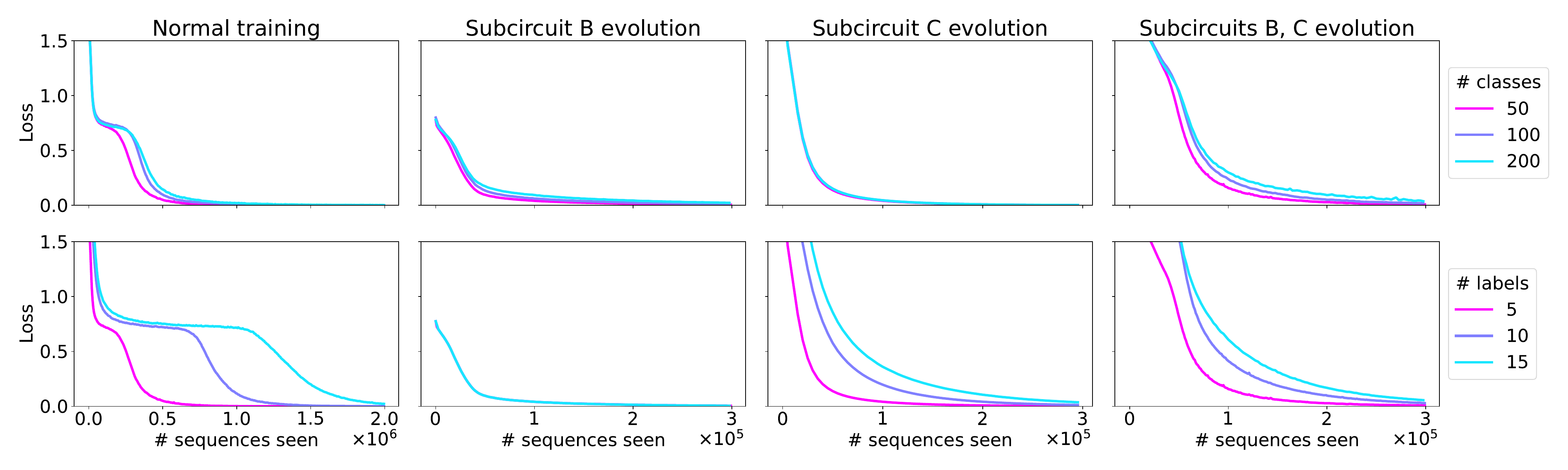}
    \vspace{-1em}
    \caption{Data-dependent learning dynamics of the induction circuit and assorted subcircuits. Top row shows curves as the \# of classes is increased, bottom row shows curves as the \# of labels is increased. Left-most column shows training loss over time without any clamps on subcircuits. Middle two columns looks at the evolution of individual subcircuits identified in Section~\ref{sec:clamping}, by clamping the other two subcircuits. Right-most column shows evolution of a composite circuit (by only clamping Subcircuit A). Middle two plots clearly show how different subcircuits depend on different data properties, and how this can explain the overall difference in learning dynamics.}
    \label{fig:data_dependent}
    \vspace{-1em}
\end{figure*}

We take this analysis further and clamp the entire match operation (green in Figures~\ref{fig:ih_schematic_methods}a, Figure~\ref{fig:subcircuits}), so only the copy operation (blue, Figure~\ref{fig:ih_schematic_methods}a) is being learned. We see this loss profile is smoothly exponential, suggesting there are no sub-components hidden within it. We then do the opposite, clamping the copy operation and focusing on the dynamics of match operation formation. We see a small saddle point followed by a quick phase change (lasting $\approx0.5e5$ iterations), indicating that the match operation should be decomposed further. We do so by clamping both layer 1 and copy components (Steps \hyperref[step:1]{1}, \hyperref[step:2]{2}, \hyperref[step:3]{3}v, \hyperref[step:5]{5}), which results in a more exponential loss profile (purple, Figure~\ref{fig:subcircuits}).

Given these results, we believe the phase change is primarily determined by three interacting subcircuits:

\textbf{Subcircuit A: Layer 1} Attending to previous token and copying it forward. Comprised of PT-attend and PT-copy (Steps \hyperref[step:1]{1} and \hyperref[step:2]{2}).

\textbf{Subcircuit B: IH QK Match} Matching queries to keys in the induction head. Comprised of Routing Q, Routing K and IH-match (Steps \hyperref[step:3]{3}qk and \hyperref[step:4]{4}).

\textbf{Subcircuit C: Copy} Copying of input label to output. Comprised of Routing V and IH-copy (Steps \hyperref[step:3]{3}v and \hyperref[step:5]{5}).

Prior work has often hypothesized the first and second components; our analysis verifies these intuitions and critically identifies the third interacting component through the use of \textit{clamping} (causal manipulations of activations throughout training).

\section{Subcircuits can explain data-dependent shifts in phase change timing}
\label{sec:data_dependent}

To demonstrate the relevance of the identified subcircuits, we turn towards explaining changes in the timing of the phase change as we modify various data properties. Previous work has shown that such data properties can affect the dynamics and emergence of ICL \cite{Chan2022, raparthy_generalization_2023, reddy2023mechanistic, yu_efficient_2023}. Specifically, we consider two possible variations: increasing the number of classes $C$ or the number of labels $L$.

In Figure \ref{fig:data_dependent}, we find that increasing both the number of classes and number of labels lead to a delay in the formation of induction heads, mostly seen as a shift in the timing of the phase change (leftmost column, Figure~\ref{fig:data_dependent}). This may be because the task becomes more challenging due to the higher data diversity. To better understand what leads to this delay, we look at the effect of these data properties on the formation of some of the subcircuits identified in Section~\ref{sec:subcircuits}, specifically Subcircuits B and C (middle two columns, Figure~\ref{fig:data_dependent}). We find that increasing the number of classes makes learning IH QK Match (Subcircuit B) harder while not increasing the difficulty of Copy (Subcircuit C), thus implying that a delay in Subcircuit B formation causes the delayed phase change. On the other hand, we find that when increasing the number of labels, IH QK Match is learnt as quickly, while copying becomes way harder, indicating that a delay in Subcircuit C formation causes the delayed phase change. These results make intuitive sense---more classes may necessitate learning a higher precision QK matching operation (Subcircuit B), and more labels may necessitate learning a higher precision copying operation (Subcircuit C). Our analysis enables this rigorous decomposition of the phase change of the full model into delays of formation of interacting subcircuits.

To emphasize the importance of considering these subcircuits individually, we consider the joint evolution of both Subcircuits B and C (which we can track by clamping Subcircuit A, analogous to the red line in Figure~\ref{fig:subcircuits}b). This would correspond to just looking at an induction head's formation, when a network is provided with previous token heads (and disentangled outputs of Layer 1). We find that both \# classes and \# labels seems to affect these dynamics, thus not providing as much clarity on how these different properties might differentially affect learning dynamics. Decomposing into the individual subcircuits that we identified in Section~\ref{sec:clamping} provides a better understanding of how data properties differentially increase the difficulty of induction circuit formation, leading to delayed phase changes.

Finally, we note that learning of the individual subcircuits is substantially faster than that of the overall dynamics. By considering data-dependent effects on individual subcircuits, we may be able to predict network behavior without having to train for long amounts of time, waiting for a phase change.

\section{Related work}
\textbf{In-context learning} From its discovery as an emergent ability in LLMs \cite{GPT3}, ICL has become a focus of study in NLP \cite{min2022rethinking, wei2023larger, wei2023symbol} and in synthetic settings \cite{vonoswald2023transformers, Chan2022, xie2021explanation}. The recent findings of transience from \citet{singh2023transient} especially motivated our work, which led to our choice of using a similar setup. Future work could extend our analyses to other setups, like linear regression or natural language tasks.

\textbf{Induction circuits} were first found by \citet{anthropicMathFramework}, then further investigated in many settings \cite{InductionHeads, wang2023label}. \citet{reddy2023mechanistic} also considers the dynamics of formation, but uses the correlational approach of \textit{progress measures}. We chose to focus on causal methods (see Appendix \ref{appx:clamp_vs_progress}) such as clamping, enabling us to discover a different set of 3 subcircuits governing induction circuit learning dynamics (Section~\ref{sec:clamping}). 

\textbf{Tooling for mechanistic interpretability} With the increasing prevalence of LLMs and corresponding safety concerns, mechanistic interpretability has seen a number of toolkits released. Primary open-source frameworks include TransformerLens \cite{nanda2022transformerlens}, \texttt{nnsight} \cite{nnsight}, and \texttt{pyvene} \cite{wu2024pyvene}. All of these frameworks are built on PyTorch, allow studying and manipulating activations from model checkpionts, and natively support many open-source LLMs. In some cases \cite{nnsight} they even allow intervention on gradients, but it's unclear if any of these frameworks can be used to manipulate activations \textit{during training with proper propagation of gradients}, which is a crucial and unique feature of our framework. Furthermore, while our library does not support as large a breadth of use cases, we hope the vibrant open-source JAX community will find a mechanistic interpretability framework based in JAX and natively supporting compilation useful (including major speedup benefits), similar to the uptake of the \texttt{Rust\_circuit} library \cite{goldowskydill2023localizing} by Rust users. The only JAX framework the authors are aware of is Tracr \cite{lindner2023tracr}, which allows researchers to quickly construct transformers that implement certain RASP programs: a complementary focus from TransformerLens, \texttt{nnsight}, \texttt{pyvene}, \texttt{Rust\_circuit} and our JAX-based artificial optogenetics framework. 

\section{Discussion}
\label{sec:discussion}
Our work analyzed induction circuits in two-layer attention-only transformers trained on a synthetic ICL task with a focus on the dynamics of formation. We took inspiration from the original discovery of induction circuits \cite{InductionHeads} and thus focused on smaller models. While a limitation of our work is the small scale compared to frontier foundation model interpretability \citep[e.g.,][]{wang2022interpretability, merullo2024circuit, palit2023visionlanguage}, we chose this setting due to academic compute constraints and a hope to dive deeper in terms of understanding. Our focus on dynamics, as well as our training setup, was inspired by recent work from \citet{singh2023transient} asserting that emergent ICL may be a dynamical phenomenon in settings where multiple competing solutions are possible to solve the task. 

Through our analysis, we discovered multiple answers to what does and doesn't need to ``go right'' for an induction head. We found that induction heads operate additively, with multiple heads used to learn the ICL task more quickly despite not being necessary to solve it. This theme of redundancy carried to previous token heads, which we found to exhibit a many-to-many wiring pattern to induction heads. Next, we used the novel \textit{clamping} methodology to identify three smoothly-evolving subcircuits whose interaction may explain the phase change in induction circuit formation. Furthermore, this better understanding helped us explain data-distribution-dependent changes in phase change timing. Of specific interest here was the quicker timelines of subcircuit formation when isolated, which could serve as useful intuition for understanding the effects various data ablations in large model training may have on the learning dynamics, without having to train full networks on each data ablation from scratch. In terms of specific subcircuits we identified, our work demonstrated the crucial role of the copy subcircuit in explaining data-distributional dependent delays in phase change timing (bottom row, Figure~\ref{fig:data_dependent}). This added understanding may have implications for practitioners when selecting a vocabulary size for LLMs: while larger vocabulary sizes are often preferred due to their increased compression ratio (and thus longer effective context), they may make copying more challenging, thereby delaying induction head formation. 

Turning back towards our original motivation from the transience of emergent ICL in transformers \cite{singh2023transient}, we hope this improved understanding of key subcircuits within induction circuits, as well as novel analysis tools, can in future shed light on why ICL may fade with overtraining.

Beyond our specific results on induction heads, our work contributes a JAX-based artificial optogenetics framework which supports compilation (yielding nearly a 50x speedup on our hardware, see Appendix~\ref{appx:cost}) and can be used to modify activations \textit{during training with proper gradient flow}, a methodology we term clamping. This framework permits causal analyses of learning dynamics, whereas prior work mostly applied causal analyses to end networks or checkpoints from normal training (see Appendix \ref{appx:clamp_vs_progress} for a comparison). We view our work as taking the first steps with this approach, and are excited to see how future work may combine our clamping methodology with other causal mechanistic interpretability techniques, such as path patching \cite{wang2022interpretability, conmy2023automated} or causal mediation analysis \cite{vig2020causal, geiger2021causal, cao2021sparse, geva2023dissecting, finlayson2021causal}, throughout training for better understanding of transformer circuit learning dynamics.

\section*{Broader impact}

This paper presents work whose goal is to advance the field of Machine Learning. There are many potential societal consequences of our work, none which we feel must be specifically highlighted here.

\section*{Acknowledgements}

We'd like to thank DJ Strouse, Kira Düsterwald, Jirko Rubruck, Andrew Lampinen, Roma Patel, Dan Roberts, Andrey Gromov, and Taylan Cemgil for useful discussions and feedback on early drafts.

A.K.S. and T.M. are funded by the Gatsby Charitable foundation. This work was supported by a Schmidt Science Polymath Award to A.S., and the Sainsbury Wellcome Centre Core Grant from Wellcome (219627/Z/19/Z) and the Gatsby Charitable Foundation (GAT3850). A.S. is a CIFAR Azrieli Global Scholar in the Learning in Machines \& Brains program.

\bibliography{references}
\bibliographystyle{icml2024}

\newpage
\appendix
\onecolumn

\section{Additional model details}
\label{appx:model}

We train causal, 2-layer attention-only transformer models, as used by prior work \cite{InductionHeads}, to study induction circuit formation. Similar to prior work \cite{Chan2022, singh2023transient}, we only train to minimize the cross entropy loss on the last token -- the predicted label for the query exemplar. We use RoPE \cite{rope} for positional encoding and a model dimension of 64, with 8 heads per layer. As is common for transformers outside of interpretability work, we are able to use LayerNorm \cite{layernorm} when reading in from the residual stream to an attention block -- LayerNorm is often excluded in interpretability work because it complicates standard analyses \cite{anthropicMathFramework}, but our causal framework works by modifying activations instead of inspecting weights or assuming linearity, circumventing the issues. 

For optimization, we use Adam \cite{adam} from the \texttt{optax} library \cite{deepmind2020jax}, with parameters $\beta_1 = 0.9, \beta_2 = 0.999$. We use a constant learning rate of $1e-5$ and a batch size of 32 sequences.

\section{Computational cost to run experiments}
\label{appx:cost}
All experiments were run on a 2018 MacBookPro with a 6-Core Intel Core i7 processor and 16GB of RAM. Standard training runs (e.g., Figure~\ref{fig:induction_heads}) took  $\approx 7$ minutes each to complete (for $1e6$ sequences seen). Clamped runs were more expensive per iteration (as some clamps require multiple forward passes), but overall took even less time since they converge in fewer sequences, note x-axis in Figure~\ref{fig:subcircuits}. This speed was possible due to the use of \texttt{jax.jit}, without which each training run would take $\approx 330$ minutes on the same hardware. Though PyTorch has recently added \texttt{torch.compile} to impart similar functionality \cite{torch_compile}, interpretability frameworks based in PyTorch are currently not compatible with \texttt{torch.compile} due to their use of hooks \cite{no_compile_transformer_lens}. Our JAX-based framework is natively compatible with \texttt{jax.jit}, as the \texttt{call\_with\_all\_aux} formalism uses PyTrees, which could possibly mean speed-ups for other interpretability researchers (as it did for us).

\section{Dataset size calculation}
\label{appx:size_calculation}

For a dataset of $C$ classes, $E$ exemplars, $L$ labels, and assuming a fraction $F$ of relabelings being used for training, our total dataset size is
\begin{align*}
    {C \choose 2} \cdot F \cdot {L \choose 2} \cdot E^3 \cdot 2 \cdot 2 \cdot 2 = 2 F E^3 C (C-1) L (L-1) \Rightarrow 1.6 C (C-1) L (L-1)
\end{align*}
for our settings where $E=1$ and $F=0.8$ are fixed.

\section{Additional details on clamping}
\label{appx:clamping}

\subsection{Implementation details of clamping for induction head formation}
\label{appx:clamp_ih_details}

In Section~\ref{sec:clamping}, we detail the methodology and inspiration for clamping various subcircuits. In this subsection, we detail the specifics of implementation of each of the clamped steps.

\subsubsection{Clamping weights vs activations: Complications due to routing}
\label{appx:clamp_routing_complicated}

\begin{figure}[ht]
    \centering
    \includegraphics[width=0.7\textwidth]{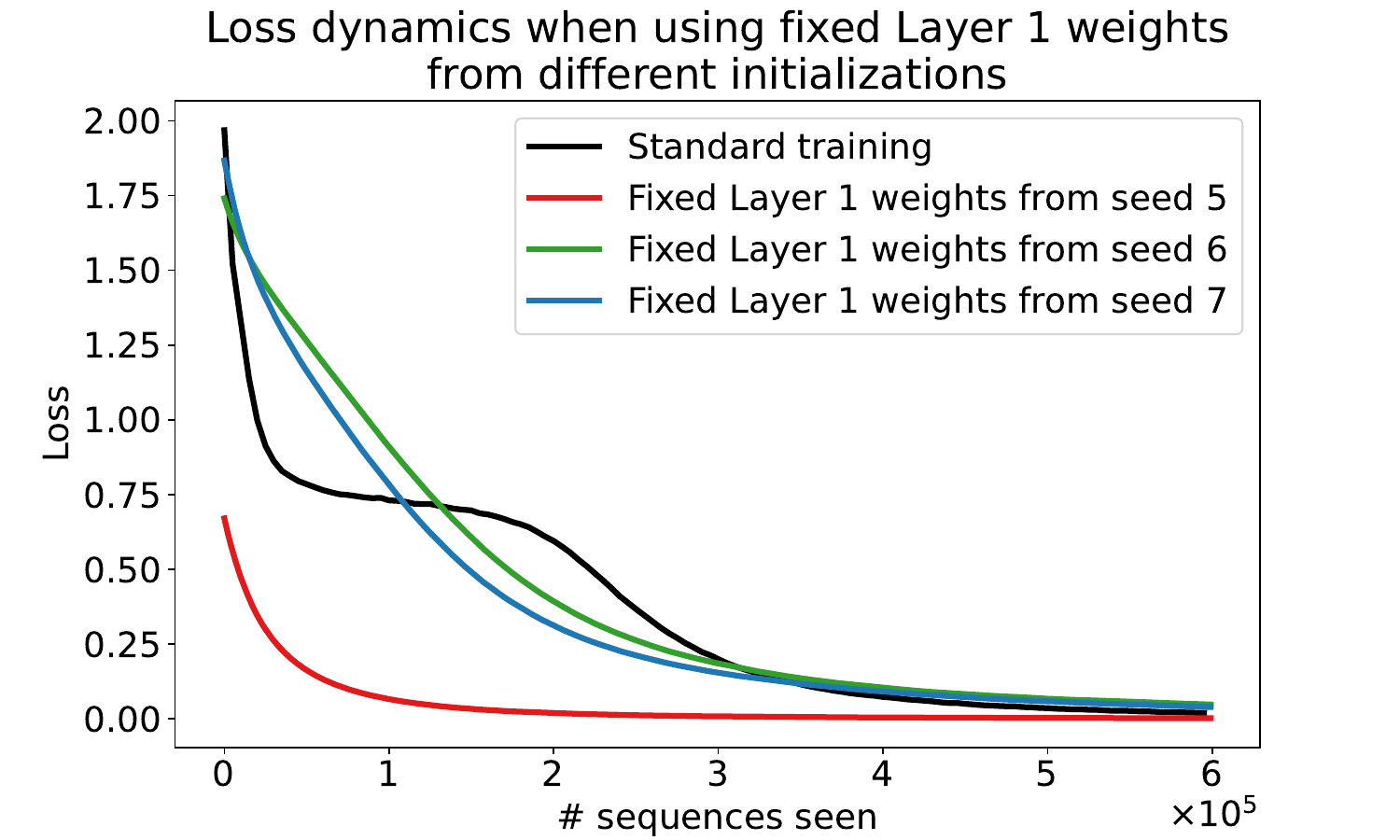}
    \caption{Loss curves when fixing Layer 1 weights to be those found at the end of training with various seeds. We see that when using the Layer 1 weights from the end of training off of the same initialization seed (seed 5) learning is way quicker, indicating that specific implementations of computations \hyperref[step:1]{1} and \hyperref[step:2]{2} in Section~\ref{sec:ih_steps} are tightly coupled to initialization. Furthermore, these results suggest that fixing weights is not the ideal form of clamping as it is very susceptible to correlations between the chosen weights and network initialization.}
    \label{fig:l1_grafts}
\end{figure}

We first demonstrate that clamping isn't as easy as just fixing weights during training. Recall that we want to clamp computation steps (Section~\ref{sec:ih_steps}) that comprise an induction circuit. There are many ways (in terms of weights) the network can implement some of these computation steps, as many of them involve reading from and/or writing into a subspace of the residual stream (e.g., Steps \hyperref[step:2]{2} and \hyperref[step:3]{3}). Arbitrary rotations could be applied to these subspaces.\footnote{This has been referred to the as \textit{identification failures} by some \cite{bactraIdentifiability}.} The specific subspaces used appear to be initialization dependent, as shown in Figure~\ref{fig:l1_grafts}.

We consider fixing the weights of Layer 1 to match those of networks pre-trained on our task, but from different initializations. Specifically, we train networks on our task from different initialization seeds (which we will refer to as seeds\footnote{These values are arbitrary and are the first three we tried. We didn't use seed 0 as this was the seed for our data sampler.} 5, 6, 7). All networks learn induction circuits, with multiple previous token heads appearing in Layer 1. We then train a network initialized from seed 5 with the residual stream after Layer 1 clamped to the residual stream after Layer 1 of one of the fully trained networks (equivalent to fixing all the weights up to this point). Note that these activations are fixed for a given input, so gradients are only flowing to Layer 2 of the model (as desired, since we are clamping Layer 1 to be an externally ``perfect'' Layer 1 based on weights). The clamped Layer 1 is theoretically perfectly performing Steps \hyperref[step:1]{1} and \hyperref[step:2]{2}. However, we can see that there is substantial variance when the weights used are from the same seed as that used for the Layer 2 initialization (red, Figure~\ref{fig:l1_grafts}) versus when the weights used are from a different seed (blue or green, Figure~\ref{fig:l1_grafts}). These results point to the importance of Step \hyperref[step:3]{3}, routing, and tell us that fixing weights may not be the best way to clamp subcircuits as it may be confounded by the specific methodology used (specifically, which network the weights come from).

\subsubsection{Activation clamps that we used}
\label{appx:clamps_used}

\begin{figure}[ht]
    \centering
    \includegraphics[width=0.7\textwidth]{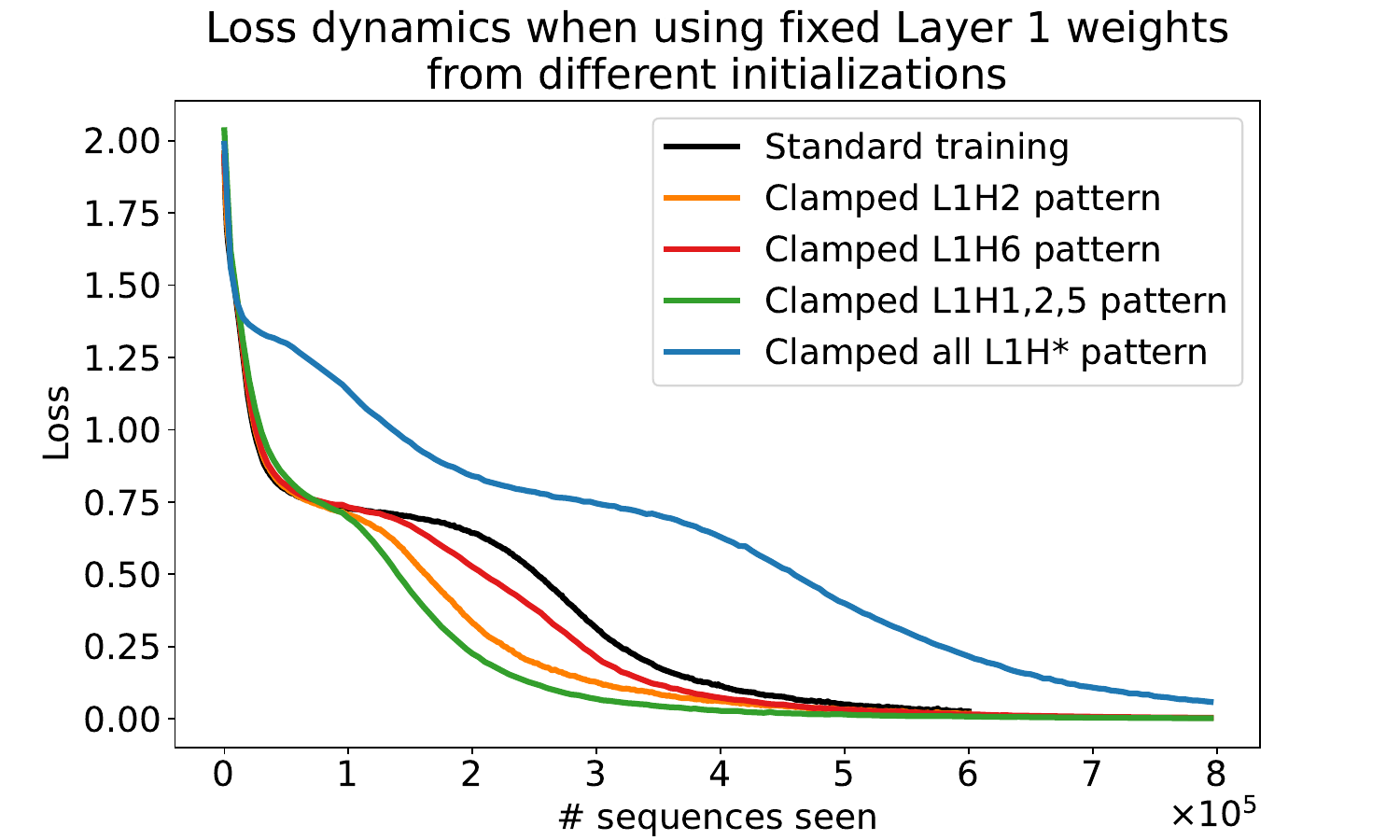}
    \caption{Effect of different implementations of the previous token attention pattern clamp (clamping computation Step \hyperref[step:1]{1}, using the terminology of Section~\ref{sec:ih_steps}). We see that most versions of the clamp have little effect on dynamics (just slight left or right shifts in the phase change), with the exception of clamping all heads in Layer 1. We suspect this may be due to some Layer 1 heads V-composing with Layer 2 heads to make copy circuits easier to learn.}
    \label{fig:pt_attend_clamps}
\end{figure}

Building off this insight, we instead implement our clamps at the activation level. While this does restrict the set of useful clamps we may apply, all the clamps used in Section~\ref{sec:clamping} can be expressed at the activation level. We detail each clamp below:

\textbf{Clamping Step \hyperref[step:1]{1} (PT-attend)} We fix the attention pattern of a Layer 1 head to always attend to its previous token (with post-softmax value 1), corresponding to clamping Step \hyperref[step:1]{1} from Section~\ref{sec:ih_steps}. In Section~\ref{sec:clamping}, we clamped Layer 1 Head 2's attention pattern. In Figure~\ref{fig:pt_attend_clamps}, we show the effect of this clamp applied to different sets of Layer 1 heads. We find that overall there is little difference, unless all heads are clamped, in which case learning is severely slowed. We hypothesize that some Layer 1 heads speed up learning of copy circuits (which is consistent with the findings of V-composition in Appendix \ref{appx:layer1}) -- by forcing all Layer 1 heads to have the previous token attention pattern, we slow this mechanism. We didn't investigate this further, however, as this was not the focus of our work, and instead used a clamp on just one previous token head for our main experiments.

\paragraph{Clamping Steps \hyperref[step:1]{1}, \hyperref[step:2]{2} (layer 1)} To implement this clamp while taking into account the concerns related to routing from Appendix \ref{appx:clamp_routing_complicated}, we consider how Layer 1 outputs are used by Layer 2. We utilize two forward passes through the network. In the first, we set the input to Layer 2 to be equivalent to the output from previous token heads and perfect query routing. For example, for the sequence ``A 0 B 1 A'', this would look like ``A A 0 B A'', where we note that the first token cannot attend to a previous token (hence it still has `A'), the next 3 tokens attend to previous tokens (so `0 B 1' becomes `A 0 B'), and the query token still routes from the input (`A'). We directly pass the embeddings of these tokens to Layer 2, and cache the \textit{patterns}. Then, we do a second forward pass using these patterns, except now we clamp the input to Layer 2 to match the input to the network (``A 0 B 1 A'') which is what the value routing should be reading from. We note that this clamp is making the task a bit easier for the network as the vectors that Routing QK needs to read from and Routing V needs to read from are not additively superimposed, but these routing steps need to learn to read from the correct subspaces. We opted for this option as it makes it even more telling that there are some subcircuit interactions in the remaining steps after applying this clamp.

\paragraph{Clamping Steps \hyperref[step:1]{1}, \hyperref[step:2]{2}, \hyperref[step:3]{3}qk, \hyperref[step:4]{4} (match)} We implement this by fixing the attention pattern of a Layer 2 head (we picked Layer 2 Head 3 as described in Section~\ref{sec:clamping}) to be that of a perfect induction head -- 1 to the correct label token and 0 to all other tokens. The aim of this clamp is to make sure the correct label is matched to (thus isolating the copy operation), which this implementation suffices to do.

\paragraph{Clamping Steps \hyperref[step:3]{3}v, \hyperref[step:5]{5} (copy)} We implement this clamp by considering the attention logits to the two labels in-context at a given induction head (we picked Layer 2 Head 3 as described in Section~\ref{sec:clamping}) and setting the output logits to them. For example, on the sequence ``A 0 B 1 A'', the attention logits from the query token `A' to the label tokens `0' and `1' would be the output logits for labels 0 and 1. The other output logits are set to $-1e9$. The aim of this clamp is to make sure labels are perfectly copied (thus isolating the match operation), which this implementation suffices to do.

\paragraph{Clamping Steps \hyperref[step:1]{1}, \hyperref[step:2]{2}, \hyperref[step:3]{3}v, \hyperref[step:5]{5}} We implement this clamp by combining the clamp for Steps \hyperref[step:1]{1}, \hyperref[step:2]{2} (layer 1), and the clamp for Steps \hyperref[step:3]{3}v, \hyperref[step:5]{5} (copy). As each of these clamps are disjoint, it's easy to apply both, serving the inteded purpose.

\subsection{Contrasting \textit{clamping} to progress measures}
\label{appx:clamp_vs_progress}

\begin{figure}[H]
    \centering
    \includegraphics[width=0.5\textwidth]{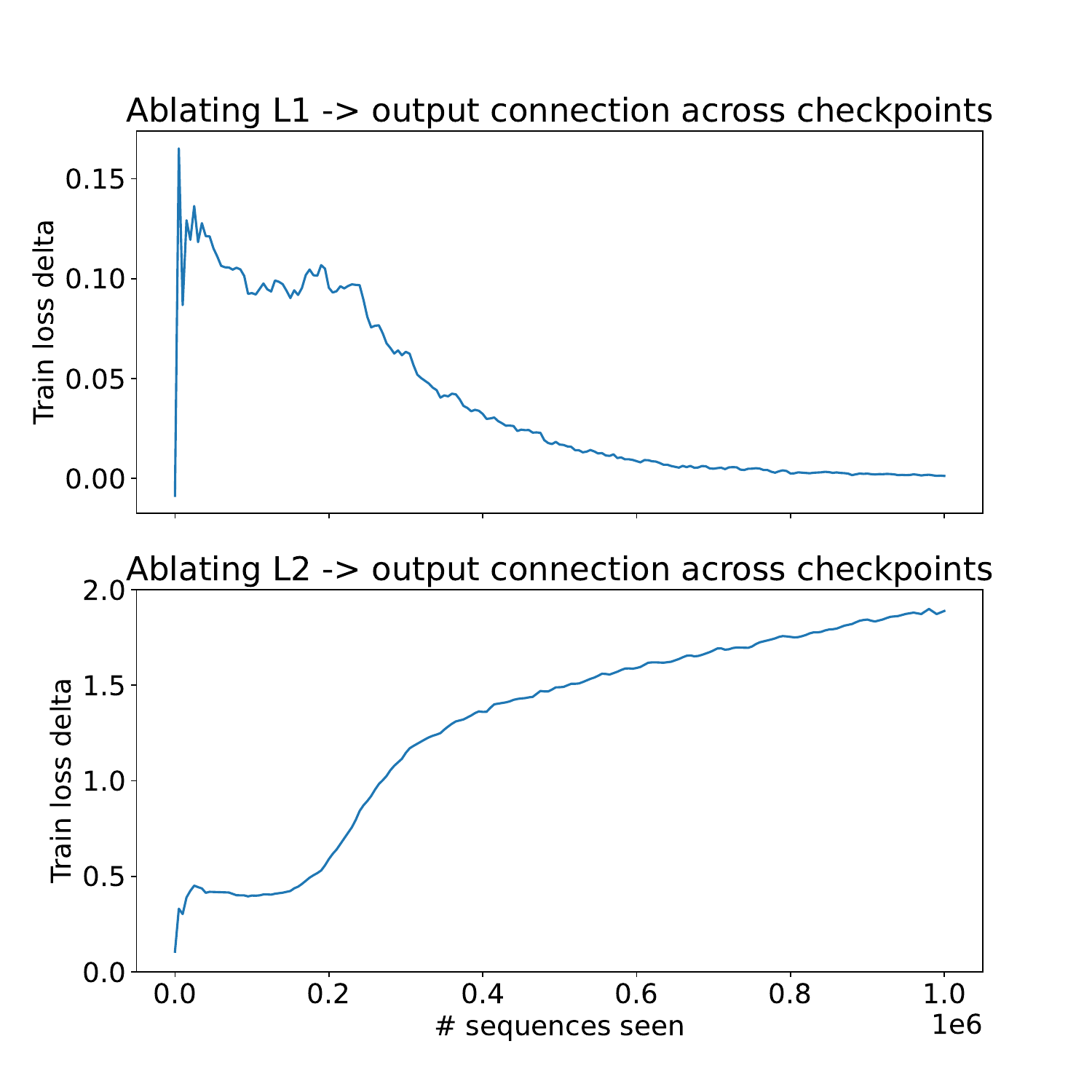}
    \vspace{-2em}
    \caption{\textbf{a)} Change in loss after ablating connection from Layer 1 heads to output (unembedding) layer. Lower change indicates that Layer 1 head to output connections are less important. The ablation was applied independently to each checkpoint from the training run shown in Figure~\ref{fig:induction_heads} and can thus be viewed as a progress measure. We can see that before the phase change, Layer 1 heads are contributing to the loss, but as induction heads form and the loss goes to 0 at the end of training, Layer 1 heads do not contribute to the output in a direct way. \textbf{b)} Same as a), but instead the connection from Layer 2 heads to output (unembedding) layer is ablated. Through training, importance of Layer 2 heads is increasing, especially after the phase change that corresponds to induction head formation.}
    \label{fig:ablate_l1_checkpoints}
\end{figure}

One prevailing current method in mechanistic interpretability for understanding training dynamics is to use \textit{progress measures}. Progress measures track quantities through the normal training process, often using causal ablations of intermediate network checkpoints.\footnote{Our results in Figure~\ref{fig:induction_heads}b and Figure~\ref{fig:ablate_l1_checkpoints} are examples of progress measures, with the latter involving a causal ablation per checkpoint.} Despite this causal nature per checkpoint, we note that progress measures are actually \textit{correlational} metrics, as the learning dynamics of the quantities they track may or may not be influencing the learning dynamics of the overall network. As such, we find our clamping method (Section~\ref{sec:clamping}) to be more useful for uncovering underlying subcircuits.

We first illustrate this with the toy model explored in Section~\ref{sec:clamping}. To track the evolution of ``subcircuit'' $\mathbf{a}$, we might consider various progress measures. The first problem we run into is that $\mathbf{a}, \mathbf{b}, \mathbf{c}$ are only specified up to scalars -- if $(\mathbf{a}, \mathbf{b}, \mathbf{c})$ is a solution, so is $(-\mathbf{a}, -\mathbf{b}, \mathbf{c})$. We find this to be an interesting toy analog to the rotations problem described in Appendix \ref{appx:clamp_routing_complicated}.  Given this, we might consider tracking the squared cosine distance between $\mathbf{a}$ and $\mathbf{a}^{true}$. Another progress measure we may consider, which offers a direct analog to clamping, is the loss if $\mathbf{b}=\mathbf{b}^{true}, \mathbf{c}=\mathbf{c}^{true}$.\footnote{Specifically, we consider $\mathbf{b}=\pm \mathbf{b}^{true}, \mathbf{c}=\pm \mathbf{c}^{true}$ and pick the one that yields lower loss (this choice is made to avoid the scaling issue).} Note the difference between this progress measure and clamping is that with the progress measure, we're considering the evolution of $\mathbf{a}$ during normal training and setting $\mathbf{b}=\mathbf{b}^{true}, \mathbf{c}=\mathbf{c}^{true}$ at every checkpoint. With clamping, we set $\mathbf{b}=\mathbf{b}^{true}, \mathbf{c}=\mathbf{c}^{true}$ throughout training and then consider the evolution of $\mathbf{a}$.

\begin{figure}
    \centering
    \includegraphics[width=\textwidth]{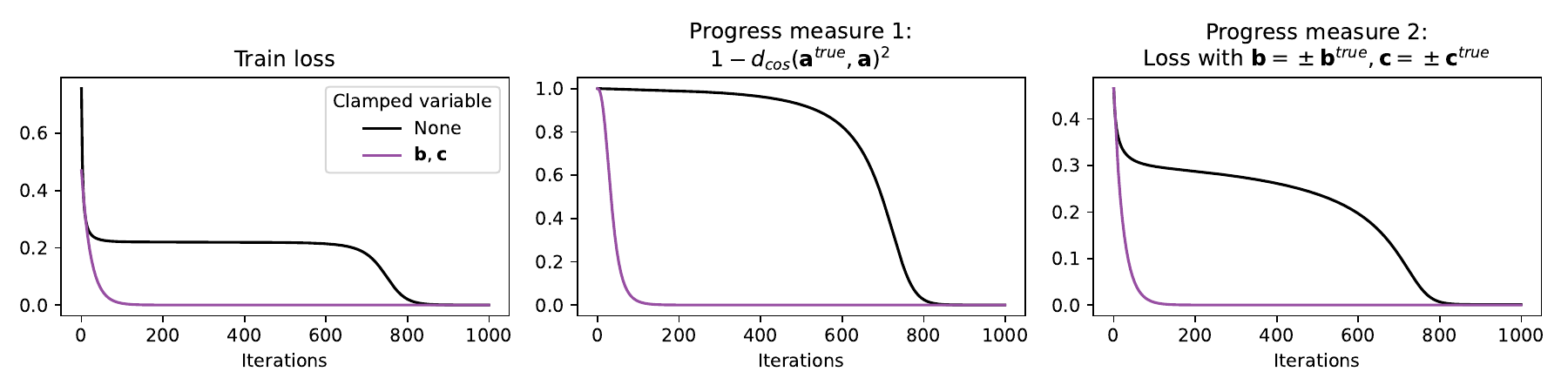}
    \caption{Progress measures vs clamping in a toy model. Black curve indicates standard training, as in Figure~\ref{fig:subcircuits}a. Leftmost plot shows the training loss. Right two plots show two different progress measures for the learning of $a$. We note that clamping (purple curve, leftmost plot) gives a clearer signal of the relevant subcircuit than progress measures in standard training (black curves, right two plots).}
    \label{fig:clamp_vs_progress_toy}
\end{figure}

Results are shown in Figure~\ref{fig:clamp_vs_progress_toy}. We find that both progress measures considered, cosine distance to true value and loss assuming $\mathbf{b}, \mathbf{c}$ fixed to those at the end of training, exhibit a phase change when observed in isolation (black curves, right two columns, Figure~\ref{fig:clamp_vs_progress_toy}). On the other hand, as seen in Figure~\ref{fig:subcircuits}a and reproduced in the leftmost plot of Figure~\ref{fig:clamp_vs_progress_toy} in purple, the loss curve when training with clamping is smoothly exponential and easily interpretable. 

\begin{figure}[h]
    \centering
    \includegraphics[width=\textwidth]{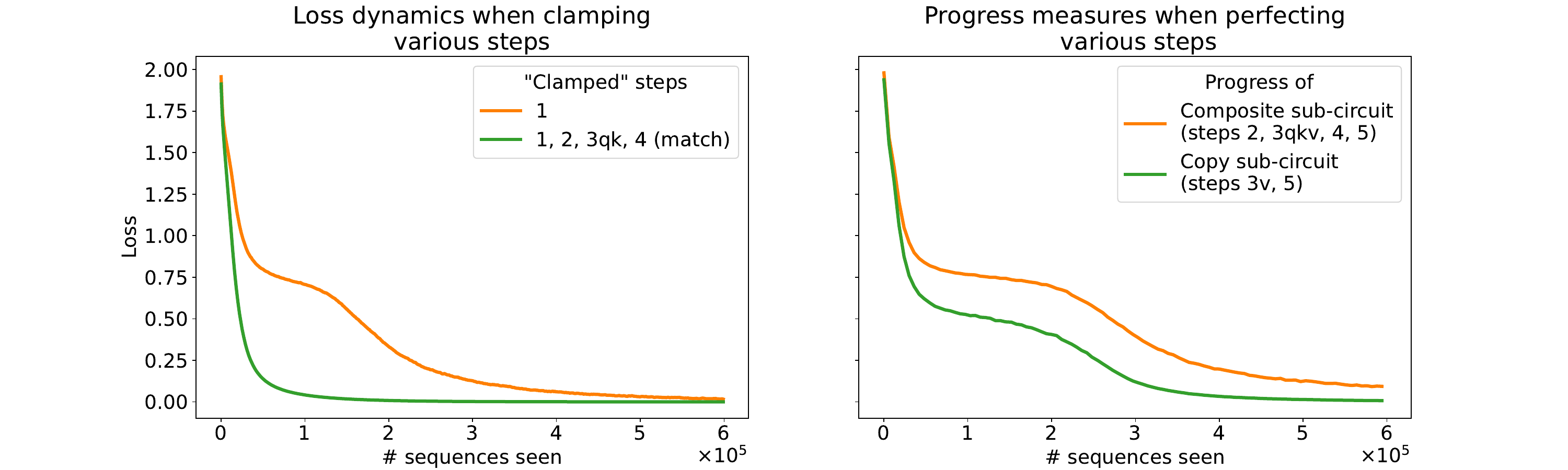}
    \caption{(Left) Our clamping method applied to studying formation of composite subcircuit consisting of Steps \hyperref[step:2]{2}, \hyperref[step:3]{3}qkv, \hyperref[step:4]{4}, \hyperref[step:5]{5} (orange) and copy subcircuit consisting of Steps \hyperref[step:3]{3}v, \hyperref[step:5]{5} (green). (Right) Progress measures tracking these same circuits. We can see that the clamping analysis more clearly shows the difference in formation dynamics of these two subcircuits, whereas the progress measure shows a phase change for both.}
    \label{fig:clamp_vs_progress}
\end{figure}

We extend results from the toy model to our setting, and consider progress measures corresponding to two of our clamped experiments: that of clamping the previous token attention pattern (clamp Step \hyperref[step:1]{1}, orange curve, Figure~\ref{fig:subcircuits}) and that of clamping the whole match operation (clamp Steps \hyperref[step:1]{1}, \hyperref[step:2]{2}, \hyperref[step:3]{3}qk, and \hyperref[step:4]{4}, green curve, Figure~\ref{fig:subcircuits}). We construct analogous progress measures by perfecting either the previous token head attention pattern or induction head attention pattern for each checkpoint in standard training and considering the loss. Figure~\ref{fig:clamp_vs_progress} shows the results, with a reproduction of the relevant curves from Figure~\ref{fig:subcircuits} for a side-by-side comparison. We can see that the progress measure is informative in the sense that it identifies the copy subcircuit (Steps \hyperref[step:3]{3}v, \hyperref[step:5]{5}) as quicker to learn than the subcircuit consisting of Steps \hyperref[step:2]{2}, \hyperref[step:3]{3}qkv, \hyperref[step:4]{4}, and \hyperref[step:5]{5}. However, when just looking at progress measures, both subcircuits appear to still emerge in a phase change, missing out on the crucial distinction between the two: the copy subcircuit (Steps \hyperref[step:3]{3}v, \hyperref[step:5]{5}) does not contain interacting components, whereas the other subcircuit (Steps \hyperref[step:2]{2}, \hyperref[step:3]{3}qkv, \hyperref[step:4]{4}, and \hyperref[step:5]{5}) does. These results show the benefit of the clamping approach for underlying exponentially forming subcircuits whose dynamics of formation causally affect the learning dynamics of the full network.

To conclude, we discuss pros and cons of progress measures versus our clamping approach. Progress measures have the benefit of only requiring access to network checkpoints after training, reducing the load of the mechanistic interpretability researcher. For cases of large language models, where training data is often not disclosed even for open-source models \cite{touvron2023llama2}, this could be especially appealing. Furthermore, as each progress measure only involves forward passes on checkpoints, they may be easier to iterate on. On the other hand, progress measures, as demonstrated above, provide largely a correlational understanding of subcircuits giving rise to dynamical phenomena. Clamping experiments, though more complex, provide a more causal understanding of learning dynamics. We hope that our results and open-source artificial optogenetics framework encourages more researchers to consider these types of experiments. 

In terms of scaling to large language models, the hope would be that isolated subcircuits evolve quickly enough that the cost is low enough. Specifically, one can view clamping experiments as requiring about 3x the cost of progress measures per iteration (due to the forwards and backwards passes). Thus, if the number of iterations needed to train clamped networks is substantially lower than that of the full network, clamping experiments may actually be cheaper than progress measures. 

\section{The role of Layer 1 heads}
\label{appx:layer1}
To establish the causal role of Layer 1 heads to individual induction heads in Layer 2, we did extensive ablation analyses.

Earlier, we found that Layer 1 heads do not substantially contribute to the output logits. To see if they only compose to Layer 2 heads' attention patterns (the match subcircuit), we consider a ``pattern preserving'' ablation \cite{InductionHeads}: the outputs of Layer 1 heads are completely ablated, but the attention patterns in Layer 2 are fixed to what they would've been had the Layer 1 heads not been ablated. We implement this using our artificial optogenetics framework by using two passes through the network. In the first, nothing is ablated. The Layer 2 attention patterns from this pass are then put into the \texttt{cache}. On the second pass through the network, Layer 1 outputs are ablated, and the \texttt{cache} is used to restore the patterns. We find that accuracy drops substantially (down to 55\%), indicating substantial V-composition in our networks. To study V-composition, we introduce an analogous ``value-preserving'' ablation where instead of caching attention patterns, Layer 2 value vectors are cached. When using just a ``value-preserving'' ablation of all Layer 1 heads (without preserving patterns), accuracy drops substantially (down to 11\%). These results indicate that Layer 1 heads are contributing to both the match subcircuit (patterns) and the copy subcircuit (values). Of the two, the role in the match circuit does seem to be ``more important'', which aligns with prior work \cite{InductionHeads}.

To verify the observed previous token (PT) heads, we next consider an ablation of these three heads with pattern-preserving (which should have no effect if they are PT heads) or value-preserving (which should have a huge effect, since patterns in Layer 2 will get messed up with these heads deleted). Likewise, we consider both types of ablations of the other heads (Layer 1 heads 0,3,4,6,7). Results are summarized in Table \ref{tab:pt_set_ablate}. We can clearly see that ablating heads 1,2,5 while preserving patterns has little effect, indicating these heads primarily compose with the match subcircuit. Likewise, we find the other heads primarily participate in V-composition for the copy subcircuit.

\begin{table}[H]
    \caption{Accuracy after Layer 1 head ablations of various sets of heads. All Layer 2 heads remain active.}
    \centering
    \begin{tabular}{lcc}
    \toprule
    \multirow{2}{*}{Ablated heads} & \multicolumn{2}{c}{Type of ablation:} \\
     & Pattern-preserving & Value-preserving \\
    \midrule
    Layer 1 Heads 1,2,5 (PT heads) & 99.85\% & 69.87\% \\
    Layer 1 Heads 0,3,4,6,7 (rest) & 34.27\% & 99.32\% \\
    \bottomrule
    \end{tabular}
    \label{tab:pt_set_ablate}
\end{table}

\begin{figure}[H]
    \centering
    \includegraphics[width=\textwidth]{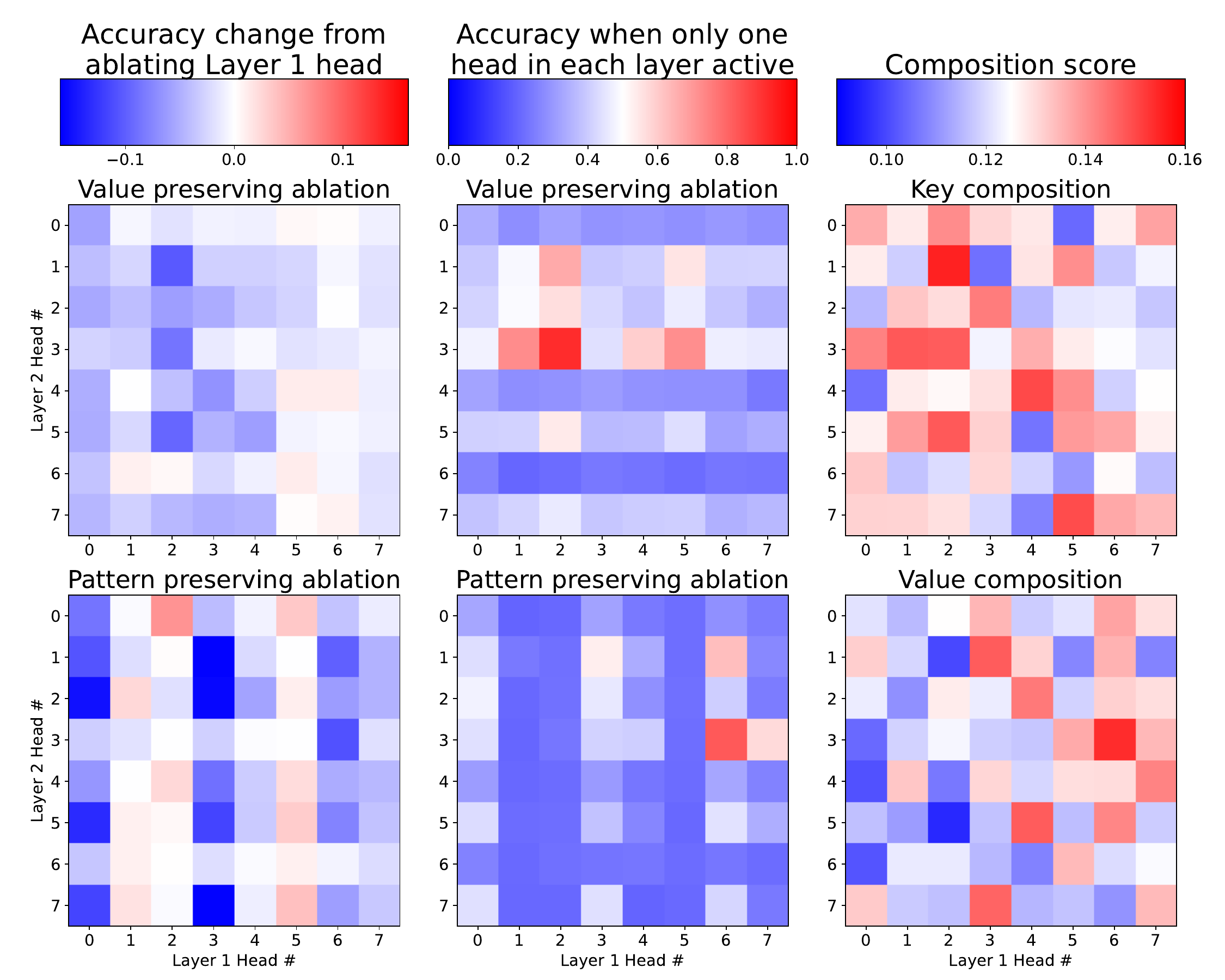}
    \vspace{-0.5em}
    \caption{Layer 1 $\rightarrow$ Layer 2 head connectivity analyses (three different methods as the three columns). \textbf{Leftmost column:} Effect of ablating a single Layer 1 head. Each column corresponds to the Layer 2 head that is kept active. In this plot, we plot the deviation compared to if the given Layer 1 head was not ablated, to emphasize positive/negative differences (here darker blue would mean more influential). Left plot preserves values, so results will be most affected by Layer 1 heads that contribute to the match subcircuit (previous token heads). From this, we see that Layer 1 Head 2 is an important previous token head for many of the Layer 2 heads. The right plot preserves patterns, so results will be least affected by Layer 1 heads that only contribute to the match subcircuit (previous token heads). In fact, we see that when ablating heads 1, 2, and 5, performance improves, indicating that previous token heads might be interfering with the network in some cases due to (suboptimal) V-composition. \textbf{Middle column:} Performance of a single pair of heads (one from Layer 1, and one from Layer 2) when preserving Layer 2 values (left) or patterns (right). In this case, we plot the raw accuracy to show how strong each path through the network is. The color scheme highlights points that go above simple context copying strategies (which has an accuracy of 50\%) in red. We see that Layer 1 Head 2 $\rightarrow$ Layer 2 Head 3 is the strongest path (when preserving values), leading to a performance of 90\%. The left plot indicates that Layer 1 heads 1, 2, and 5 play a strong role in computing patterns. The right plot indicates that Layer 1 Head 6 has the most important V-composition to Layer 2 heads. \textbf{Rightmost column:} For comparison, we include the composition score \cite{nanda_dynalist, anthropicMathFramework} of Layer 1 head output projections to key and value input matrices in Layer 2. We again use a bwr colorscheme to highlight pairs with higher than average composition scores. While the main forms of composition are still visible (K-composition from Layer 1 Head 2 to Layer 2 Head 1/3, and V-composition from Layer 1 Head 6 to Layer 2 Head 3), we note that the results are overall way noisier. This suggests that the activation modification approach used in the middle column, made easy to implement by our artificial optogenetics framework, may provide clearer signals than weight-based approaches on circuit wiring. 
    }
    \label{fig:pt_indiv_contrib}
\end{figure}

Finally, we wish to establish the connectivity between Layer 1 and Layer 2 heads. To be specific to Layer 2 head, we repeat each experiment 8 times, once for each Layer 2 head active (and the others are ablated). As in Section~\ref{sec:additive_heads}, we first consider ablating a single previous token head. Results are shown in Figure~\ref{fig:pt_indiv_contrib}a. We find that when ablating single heads, only Head 2 seems to have a strong effect. We believe this result is again suffering from the redundancy across heads, so the importance of heads 1 and 5 is masked. To account for this, we instead consider ablating all but a given Layer 1 head. In this case, each square in the heatmap can be seen as one path through the network, with only one head in each layer used. Results are shown in Figure~\ref{fig:pt_indiv_contrib}b. From these plots, it's clear that there are multiple active paths in the network: 
\begin{enumerate}
    \item Layer 1 Head 1 $\rightarrow$ Layer 2 Head 3
    \item Layer 1 Head 2 $\rightarrow$ Layer 2 Heads 1, 2, 3
    \item Layer 1 Head 5 $\rightarrow$ Layer 2 Heads 1 and 3
\end{enumerate}
These results indicate a \textit{many-to-many wiring diagram}.

We'd also like to take a moment here to emphasize the use of activation-level ablation analyses, as opposed to metrics calculated on weights. Composition score \cite{nanda_dynalist, anthropicMathFramework} is one such weight-based metric, calculated as:
\begin{align*}
    \text{score}\left(L_1H_i \rightarrow L_2H_j : Q/K/V\right) = \frac{||W^{L_1H_i}_O W^{L_2H_j}_{Q/K/V}||_F}{||W^{L_1H_i}_O||_F ||W^{L_2H_j}_{Q/K/V}||_F}
\end{align*}
It aims to measure the connection strength between heads in different layers by considering the matrices they use to write output and read input from the residual stream. In Figure~\ref{fig:pt_indiv_contrib}c, we plot the composition scores of Layer 1 heads to key and value matrices of Layer 2 heads. Though the most noteworthy cases of composition are recovered (such as the importance of Layer 1 Head 2 for K-composition, and Layer 1 Head 6 for V-composition), overall we find the results to be much harder to interpret than those using our activation-level ablation analyses (Figure~\ref{fig:pt_indiv_contrib}b).

\clearpage

\section{Proof of saddle point in toy model}
\label{appx:saddle_point}

We include a brief proof that $\mathbf{a} = \mathbf{b} = \mathbf{c} = \mathbf{0}$ is a saddle point for the toy model, leading to the long loss plateau before the phase change.

Recall, the loss being minimized is:
\begin{align*}
    \mathcal{L}(\mathbf{a}, \mathbf{b}, \mathbf{c}) = \frac{1}{2}\sum_{i,j,k}\left(a^{true}_i b^{true}_j c^{true}_k - a_i b_j c_k\right)^2,
\end{align*}
and we assume $\mathbf{a}^{true},\mathbf{b}^{true},\mathbf{c}^{true} \neq \mathbf{0}$.

We then have:
\begin{align*}
    \frac{\partial \mathcal{L}(\mathbf{a}, \mathbf{b}, \mathbf{c})}{\partial a_i} &= - \sum_{j,k}\left(a^{true}_i b^{true}_j c^{true}_k - a_i b_j c_k\right) b_j c_k = 0, \\
    \frac{\partial \mathcal{L}(\mathbf{a}, \mathbf{b}, \mathbf{c})}{\partial b_j} &= - \sum_{i,k}\left(a^{true}_i b^{true}_j c^{true}_k - a_i b_j c_k\right) a_i c_k = 0, \\
    \frac{\partial \mathcal{L}(\mathbf{a}, \mathbf{b}, \mathbf{c})}{\partial c_k} &= - \sum_{i,j}\left(a^{true}_i b^{true}_j c^{true}_k - a_i b_j c_k\right) a_i b_j = 0,
\end{align*}
if $\mathbf{a} = \mathbf{b} = \mathbf{c} = \mathbf{0}$. Moving forward, we will abbreviate derivations using the symmetry between $a, b, c$.

The value of the loss function at this point is:
\begin{align*}
    \mathcal{L}(\mathbf{0}) = \frac{1}{2}\sum_{i,j,k}\left(a^{true}_i b^{true}_j c^{true}_k\right)^2
\end{align*}

While we know this point is neither a global maximum ($\infty$) nor a global minimum (0) of the loss function, we now need to show it's a saddle point (and not a local minimum/maximum).

\subsection{Hessian test inconclusive for 3 or more interacting variables}
A first attempt to show it's a saddle point (and not a local minimum/maximum) would be to use the determinant of the Hessian. We note that such a proof does work in the two vector case (the determinant of the Hessian is negative in that case), but for the 3-variable case, the determinant of the Hessian is also 0. We show this below:

The Hessian can be seen as a $3\times 3$ block matrix:
\begin{align*}
    \mathcal{H} = \begin{pmatrix}
        H_{aa} & H_{ab} & H_{ac} \\
        H_{ba} & H_{bb} & H_{bc} \\
        H_{ca} & H_{cb} & H_{cc} \\
    \end{pmatrix},
\end{align*}
and we can work out the individual terms below:
\begin{align*}
    \frac{\partial^2 \mathcal{L}(a, b, c)}{\partial a_i^2} &= \sum_{j,k} (b_j c_k)^2 = 0 \\
    \frac{\partial^2 \mathcal{L}(a, b, c)}{\partial a_i \partial a_{i'}} &= 0, i \neq i'
\end{align*}
so $H_{aa} = H_{bb} = H_{cc} = \mathbf{0}$. For the cross terms, we have:
\begin{align*}
    \frac{\partial^2 \mathcal{L}(a, b, c)}{\partial a_i \partial b_j} = \sum_{k} 2 a_i b_j c_k^2 - a^{true}_i b^{true}_j c^{true}_k c_k = 0
\end{align*}

Here, the presence of $c_k$ forces the these partials to 0, making the Hessian a matrix of all 0's. Note in the two variable case (if we remove $\mathbf{c}$), the Hessian would have positive off-diagonal elements and it becomes easy to show the determinant is negative.

\subsection{Directions of positive and negative slope}
Since the Hessian test doesn't work, we can instead attempt to identify two directions, one in which the function will increase in a small neighborhood and one in which it will decrease in a small neighborhood. These are easy to construct by picking $i', j', k'$ s.t. $a_{i'}^{true}b_{i'}^{true}c_{i'}^{true} \neq 0$ (such indices are guaranteed to exist since $\mathbf{a}^{true},\mathbf{b}^{true},\mathbf{c}^{true} \neq \mathbf{0}$).

Specifically, consider the direction of $\mathbf{d_-} =  \langle 0, ..., 0, a^{true}_{i'}, 0, ..., 0, b^{true}_{j'}, 0, ..., 0, c^{true}_{k'}, 0, ..., 0\rangle$. We can see that a small $\varepsilon > 0$\footnote{Note, the positivity assumption is not necessary. In fact, if we allow $\varepsilon$ to take either sign, we can just use  one direction to see the saddle point (due to the factor of $\varepsilon^3$). However, for the case of an even number of components (e.g., $N=4$ instead of $N=3$), we do need two separate directions (since $\varepsilon^N$ will always be positive if $N$ is even). To make this proof general, we thus provide the two directions and focus on positive $\varepsilon$.} perturbation along this direction will reduce the loss:
\begin{align*}
    \mathcal{L}(\mathbf{0}+\varepsilon  \mathbf{d_-}) &= \frac{1}{2}\sum_{i,j,k}\left(a^{true}_i b^{true}_j c^{true}_k - a_i b_j c_k\right)^2 \\
    &= \frac{1}{2} \left(a^{true}_{i'} b^{true}_{j'} c^{true}_{k'} - \varepsilon^3 a^{true}_{i'} b^{true}_{j'} c^{true}_{k'}\right)^2 + \frac{1}{2}\sum_{i\neq i',j\neq j',k \neq k'}\left(a^{true}_i b^{true}_j c^{true}_k\right)^2 \\
    &= \frac{1}{2} \left(a^{true}_{i'} b^{true}_{j'} c^{true}_{k'}\right)^2 (\varepsilon^6 -2\varepsilon^3) +\frac{1}{2}\sum_{i,j,k}\left(a^{true}_i b^{true}_j c^{true}_k\right)^2 \\
    \Rightarrow \lim_{\varepsilon \rightarrow 0} \mathcal{L}(\mathbf{0}+\varepsilon  \mathbf{d_-}) - \mathcal{L}(\mathbf{0}) &= \frac{1}{2} \left(a^{true}_{i'} b^{true}_{j'} c^{true}_{k'}\right)^2 (\varepsilon^6 -2\varepsilon^3) \approx - \varepsilon^3 \left(a^{true}_{i'} b^{true}_{j'} c^{true}_{k'}\right)^2 < 0
\end{align*}
Thus, a direction with a negative directional derivative exists.

Similarly, we consider $\mathbf{d_+} =  \langle 0, ..., 0, -a^{true}_{i'}, 0, ..., 0, b^{true}_{j'}, 0, ..., 0, c^{true}_{k'}, 0, ..., 0\rangle$. We can see that a small $\varepsilon > 0$ perturbation along this direction will increase the loss:
\begin{align*}
    \mathcal{L}(\mathbf{0}+\varepsilon  \mathbf{d_+}) &= \frac{1}{2}\sum_{i,j,k}\left(a^{true}_i b^{true}_j c^{true}_k - a_i b_j c_k\right)^2 \\
    &= \frac{1}{2} \left(a^{true}_{i'} b^{true}_{j'} c^{true}_{k'} + \varepsilon^3 a^{true}_{i'} b^{true}_{j'} c^{true}_{k'}\right)^2 + \frac{1}{2}\sum_{i\neq i',j\neq j',k \neq k'}\left(a^{true}_i b^{true}_j c^{true}_k\right)^2 \\
    &= \frac{1}{2} \left(a^{true}_{i'} b^{true}_{j'} c^{true}_{k'}\right)^2 (\varepsilon^6 +2\varepsilon^3) +\frac{1}{2}\sum_{i,j,k}\left(a^{true}_i b^{true}_j c^{true}_k\right)^2 \\
    \Rightarrow \lim_{\varepsilon \rightarrow 0} \mathcal{L}(\mathbf{0}+\varepsilon  \mathbf{d_+}) - \mathcal{L}(\mathbf{0}) &= \frac{1}{2} \left(a^{true}_{i'} b^{true}_{j'} c^{true}_{k'}\right)^2 (\varepsilon^6 +2\varepsilon^3) \approx \varepsilon^3 \left(a^{true}_{i'} b^{true}_{j'} c^{true}_{k'}\right)^2 > 0
\end{align*}
Thus, we've identified two directions, one which increases the loss and one which decreases the loss, indicating that $\mathbf{a} = \mathbf{b} = \mathbf{c} = \mathbf{0}$ must be a saddle point. Note, if we find the directional derivative in either of these directions, it's still 0 (due to the factor of $\varepsilon^3$ not vanishing, even if we divide by $\varepsilon$ to get the derivative)---we simply show the derivation to show the differing sign. We also note that this argument extends beyond 3 interacting vectors.

This argument can also give some intuition as to why the saddle point is ``harder to escape'' in the case of 3 interacting vectors (as evidenced by the longer loss plateau in the black curve versus the blue curve in Figure~\ref{fig:subcircuits}a). Here, we have a factor of $\varepsilon^3$ rather than $\varepsilon^2$ (in the two vector case). The intuition extends: if the number of components is increased further (beyond 3), the saddle point will get harder and harder to escape.

\section{Assorted supplementary figures}
\label{appx:extra_figures}

While writing the main paper, we tried to maintain a coherent narrative that highlights the main results and intuitions we want readers to take away. For the in-the-trenches researcher, we provide some additional results that we thought were interesting.

\begin{figure}[H]
    \centering
    \includegraphics[width=0.5\textwidth]{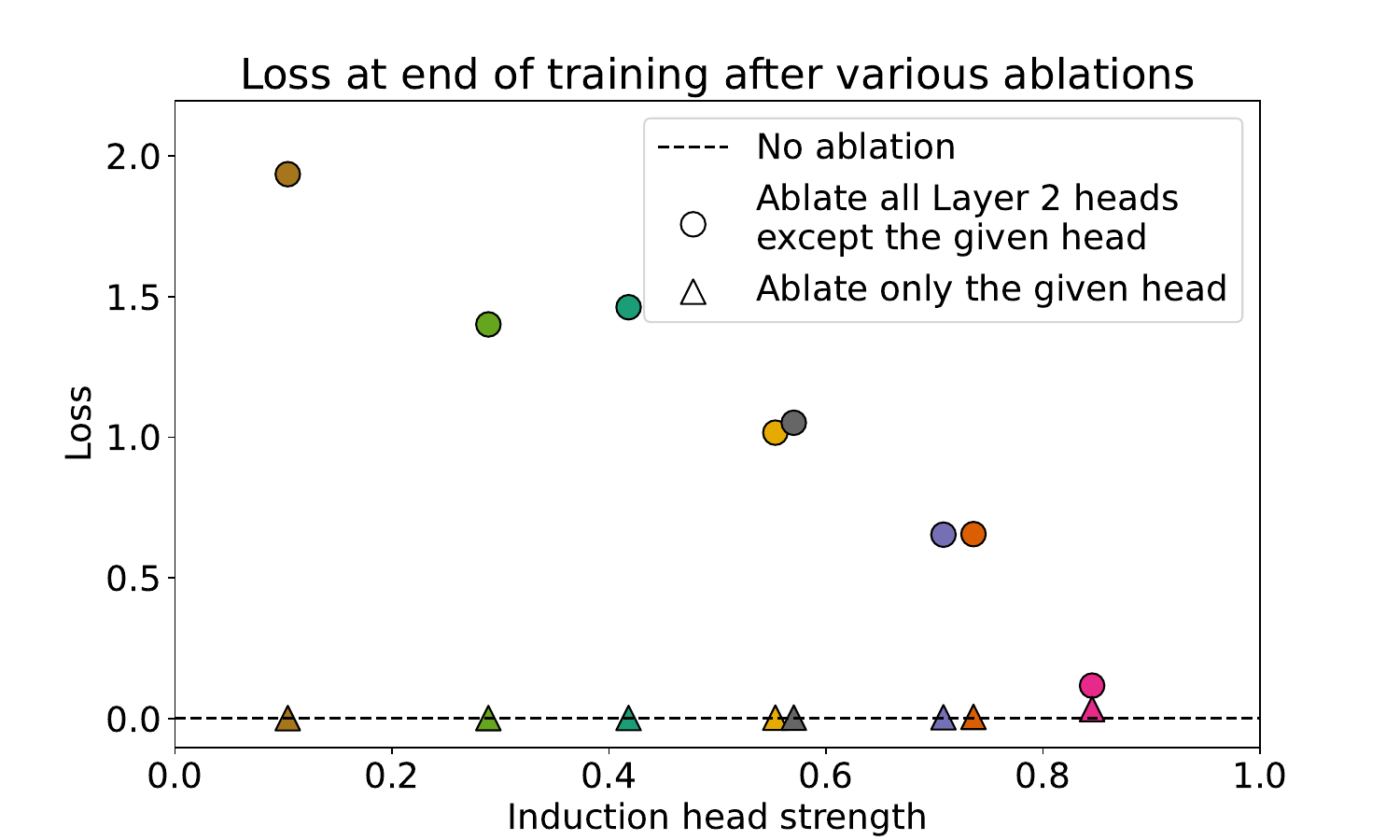}
    \caption{Same as Figure~\ref{fig:additive_heads}a, but plotting loss instead of accuracy. We include this plot for completeness, and used the accuracy version as we felt it was clearer for the main text.}
    \label{fig:additive_heads_loss}
\end{figure}

\begin{figure}[H]
    \centering
    \includegraphics[width=\textwidth]{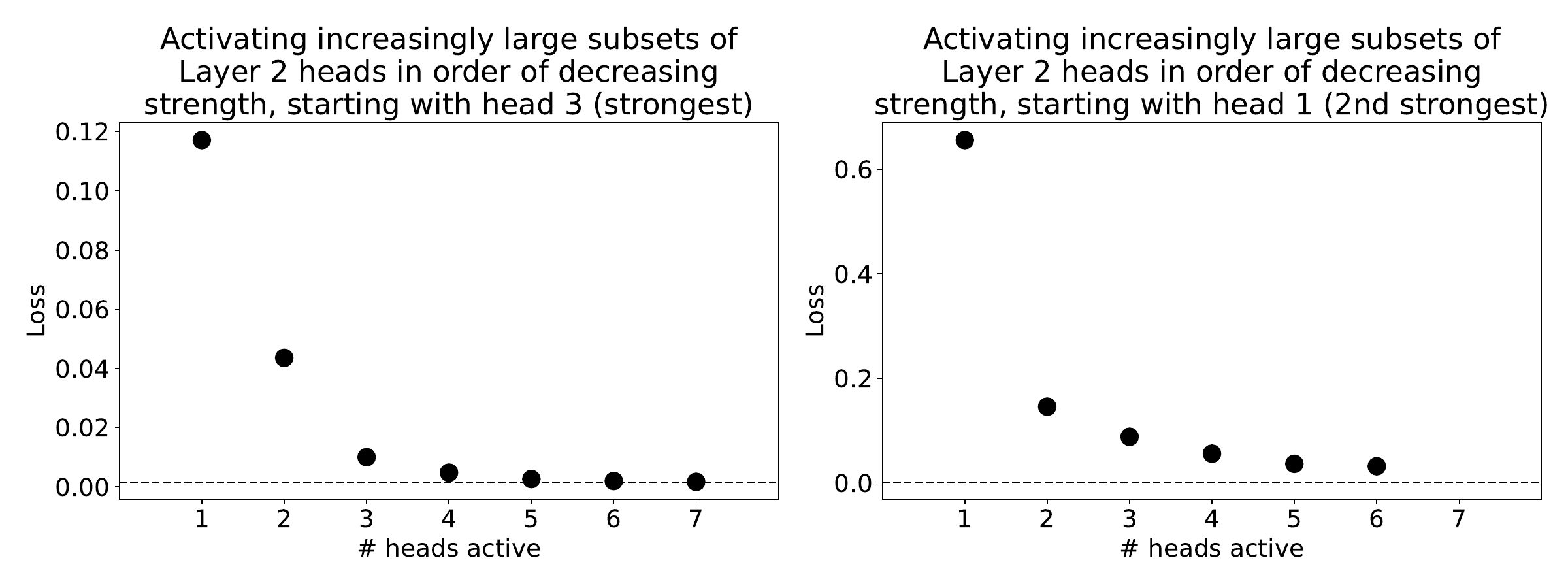}
    \vspace{-2em}
    \caption{Another way of seeing the additive effect of induction heads. We first order the heads by strength (this happens to be \texttt{[3,1,2,7,5,0,4,6]} in our case, see Figure~\ref{fig:induction_heads}b). Then, we consider an ablation where only a certain number of heads (x-axis) are left active. In the left plot, we start with Head 3, so the points from left to right indicate [only Head 3 active], [Heads 3 and 1 active], [Heads 3 and 1 and 2 active], and so on. The right plot is the same, except we never activate Head 3 (the strongest head). The monotonically decreasing loss in both plots indicate that heads have an additive effect on each other. Dotted line indicates loss of full network (no ablations). Note: the y-axis on the right is scaled larger since Head 3 (the strongest head) is never active (so losses are higher).}
    \label{fig:additive_heads_one_at_a_time}
\end{figure}

\begin{figure}[H]
    \centering
    \includegraphics[width=\textwidth]{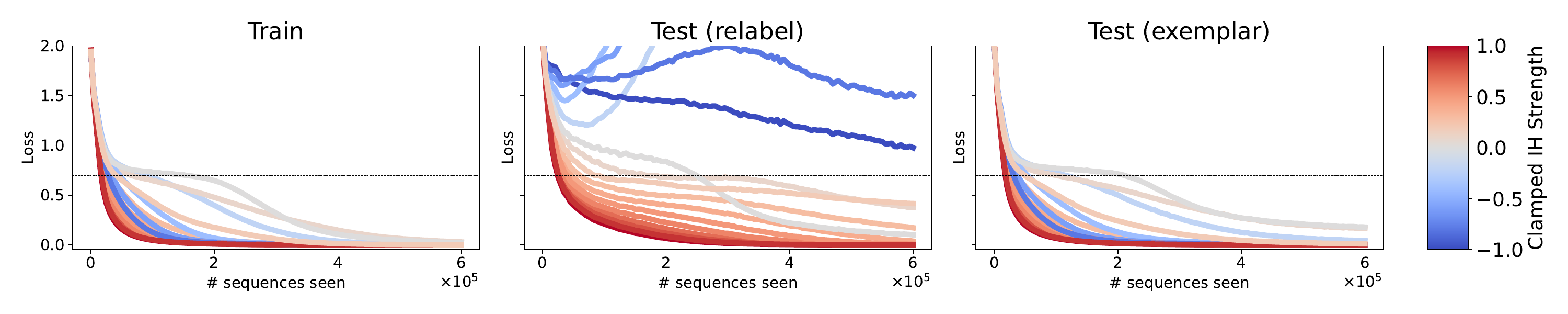}
    \vspace{-2em}
    \caption{Clamping through training with an imperfect match circuit. For this figure, we clamped the attention pattern of a single Layer 2 head to look like a noisy induction head (the other heads were left active) to see how ``strong'' an induction head needs to be before it induces the exponential learning dynamics seen when clamping to maximum strength (green line, Figure \ref{fig:subcircuits}). For example, when ``Clamped IH strength'' is 0.4, this corresponds to attention of 0.7 to the correct label token and 0.3 to the incorrect label token, and 0 to all other tokens. We vary the noisiness in the induction head and see how this effects dynamics. The darkest red curve (perfect match) corresponds to just learning a copy subcircuit (as in Figure \ref{fig:subcircuits}, green line), exhibiting a smooth exponential decay pattern. As the strength is turned down to 0, we see the learning dynamics shift from exponential to slow exponential to the usual  phase change when the IH strength is clamped to 0 (as this head is ignored, and the network learns using the other Layer 2 heads). When we fix the induction strength to -1, meaning the head is \textit{only} paying attention to the wrong token, we recover an exponential curve. This is an example of the network using an ``anti-induction'' head (a phenomenon we sometimes observed if $L=2$): the mechanism here is the network is \textit{copying} both labels from context and then \textit{subtracting out} the wrong label. This is an example of a slightly different (and perhaps more complicated) circuit for implementing ICL in our task, similar to how \cite{zhong2023clock} discovered two different circuits for modular addition. We hypothesize that the reason this ``anti-induction'' circuit doesn't emerge naturally when $L > 2$ is because imperfect versions of ``anti-induction'' heads (e.g., ``Clamped IH strength'' $=-0.8$) are less robust to different label pairs (middle plot, light-dark blue curves), which may make it a circuit element that is feasible but not learnable (under this task setup).}
    \label{fig:enter-label}
\end{figure}

\section{The ``residual heads'' hypothesis}
\label{sec:residual_heads}

In Section~\ref{sec:additive_heads}, we provide evidence that induction heads function additively with redundancy across heads. Combining this insight with the ordering of head emergence in Figure~\ref{fig:induction_heads}b, we hypothesize that other induction heads form to minimize the residual of the loss left after earlier heads form (e.g. Head 3). We didn't include this hypothesis in the main paper as it's not the focus of our work, and we find evidence supporting and opposing it. As scientists, we hope these mixed results can serve as a launch point for further study, and thus choose to include them here.

One simple way to test this hypothesis would be to consider the mistakes made by the individual Layer 2 heads, when all other Layer 2 heads are turned off. Let's consider Head 2 and Head 3 for example. Head 3 on its own achieves an accuracy of 98\%, outputting the wrong answer for 1553 of the 78400 training sequences. Head 2's accuracy is 75\%. If Head 2 is fitting to the residual, we might expect Head 2's accuracy on the sequences that Head 3 fails to solve to be above its baseline of 75\%, meaning it's preferentially stronger on the problems where Head 3 is mistaken. On the other hand, it could be that the sequences missed by Head 3 are simply ``harder'', in which case we would expect an accuracy less than its baseline. In Table \ref{tab:residual_head_test}, we find that the latter is  more likely to be the case, so this piece of evidence is \textit{against} the residual heads hypothesis.

\begin{table}[h]
    \vspace{-1em}
    \caption{Accuracy when ablating all but a given Layer 2 head on the subset of problems that are incorrect when ablating all but Layer 2 Head 3 (the one with highest induction score). Second row contains the accuracies over the whole dataset for comparison (which matches the circles in Figure~\ref{fig:additive_heads}a).}
    \centering
    \begin{tabular}{lccccccc}
    \toprule
    Accuracy & Head 0 & Head 1 & Head 2 & Head 4 & Head 5 & Head 6 & Head 7 \\
    \midrule
    On the subset & 24.08\% & 64.78\% & 71.67\% & 32.97\% & 56.41\% & 16.61\% & 46.10\% \\
    Overall & 40.32\% &  76.51\% & 75.12\% & 42.85\% & 64.65\% & 29.16\% & 58.85\% \\
    \bottomrule
    \end{tabular}
    \label{tab:residual_head_test}
\end{table} 

\begin{table}
    \caption{Same as Table \ref{tab:residual_head_test}, except we also ``perfect'' each induction head's match circuit (as done in Figure~\ref{fig:ih_imperfect_copier}), so we can view this table as only considering the mistakes in each induction head's copy circuit. There are only 58 examples that the ``perfected match'' layer 2 Head 3 gets wrong, so this table only considers the accuracy of other ``perfected match'' heads on these 58 examples. Note the overall accuracies in this Table match the squares in Figure~\ref{fig:ih_imperfect_copier}.}
    \centering
    \begin{tabular}{lccccccc}
    \toprule
    Accuracy & Head 0 & \textbf{Head 1} & Head 2 & Head 4 & Head 5 & Head 6 & Head 7 \\
    \midrule
    On the subset & 24.14\% & \textbf{86.21\%} & 20.69\% & 22.41\% & 50.00\% & 0.00\% & 44.83\% \\
    Overall & 50.01\% &  82.80\% & 82.73\% & 62.10\% & 65.42\% & 28.46\% & 70.70\% \\
    \bottomrule
    \end{tabular}
    \label{tab:residual_head_test_perfect_copy}
\end{table} 

\begin{figure}[t]
    \centering
    a)\includegraphics[width=0.47\textwidth]{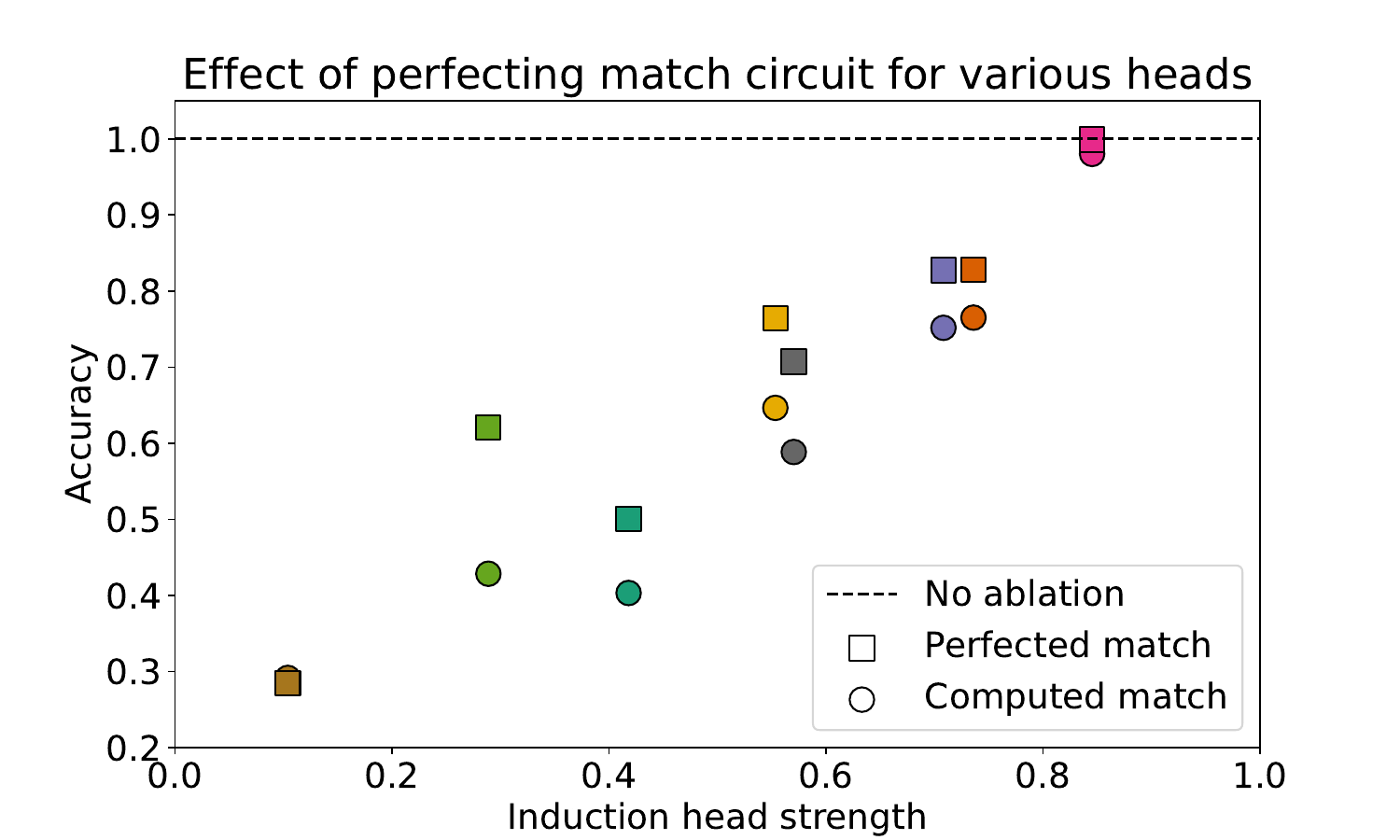}\\
    b)
    \includegraphics[width=0.47\textwidth]{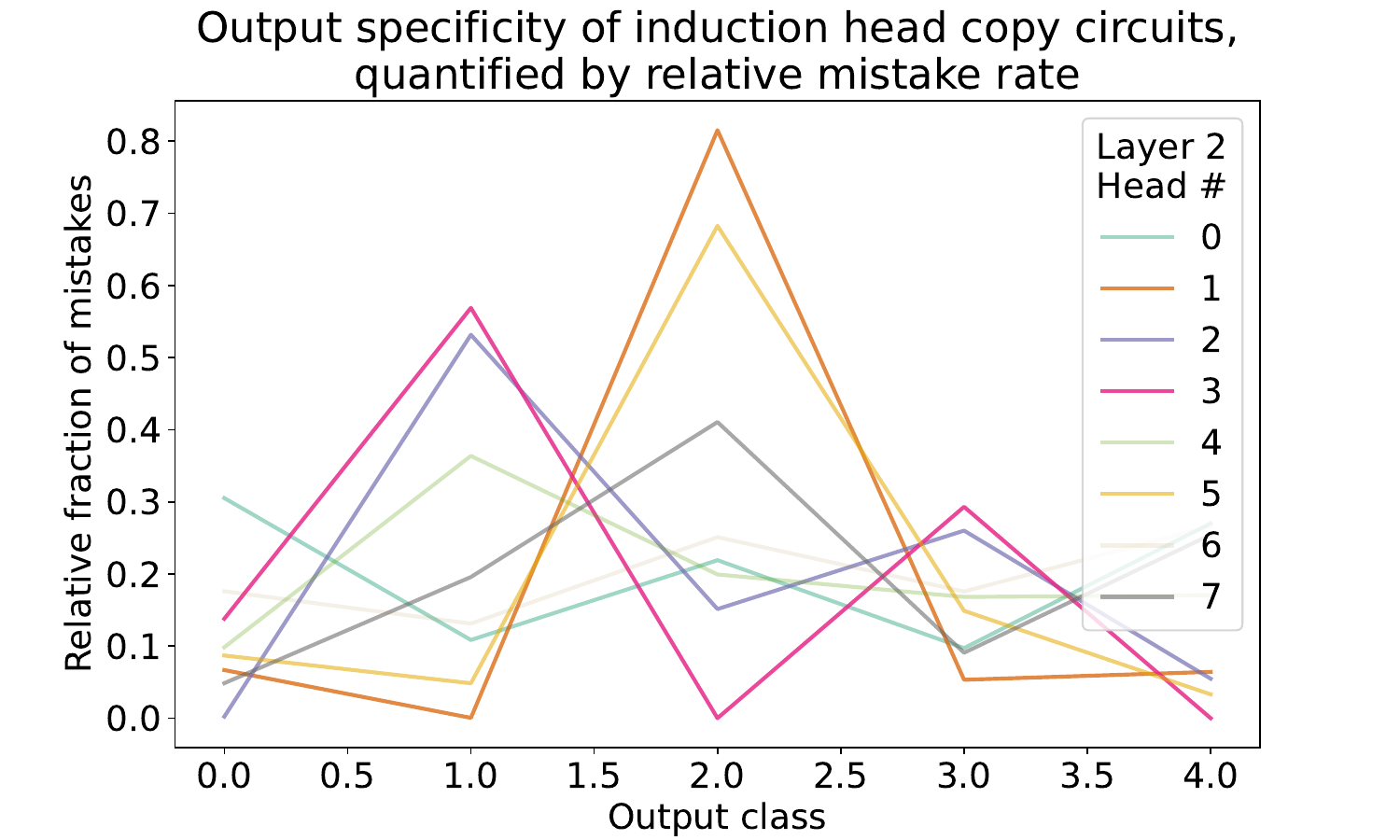}
    c)\includegraphics[width=0.47\textwidth]{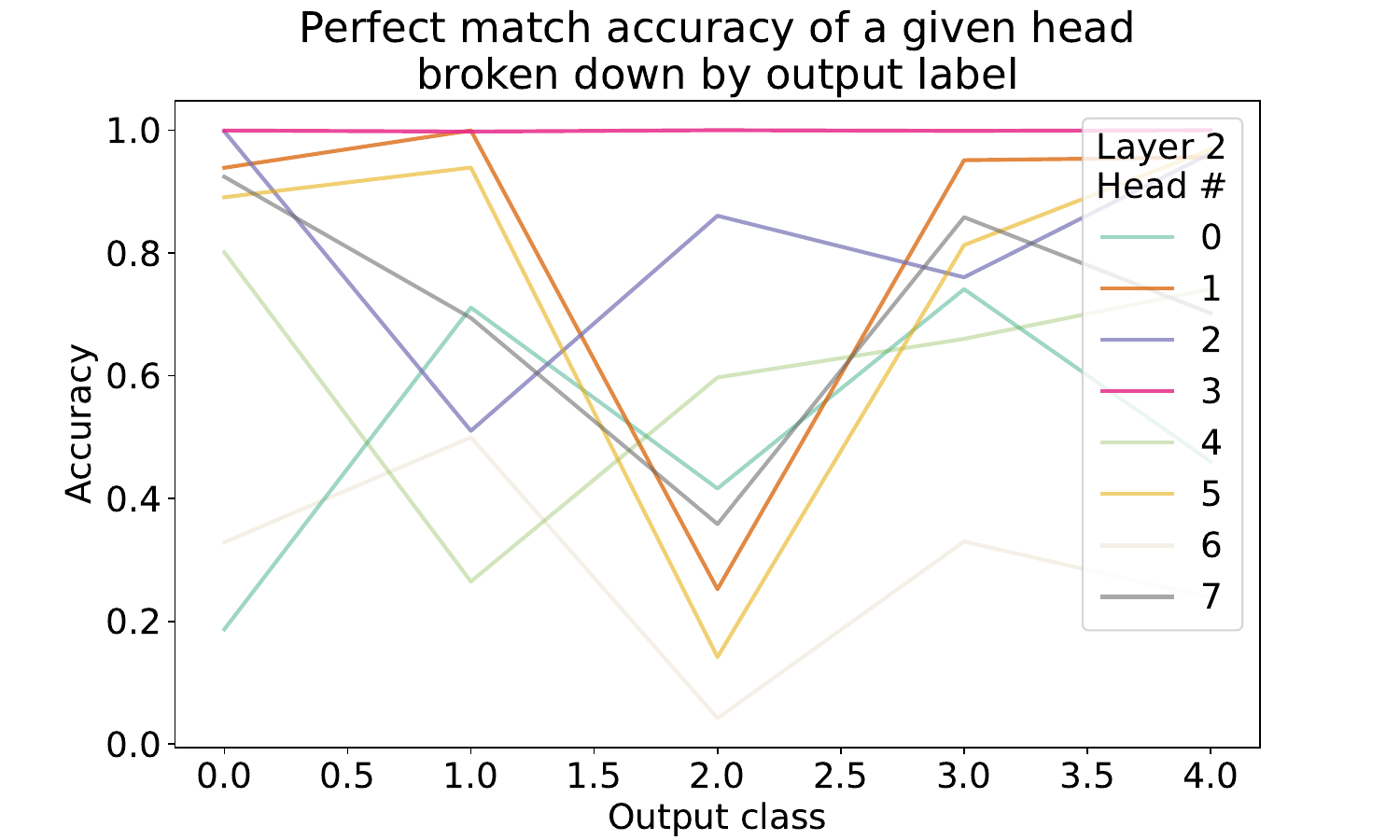}
    \caption{\textbf{a)} Induction heads are imperfect copiers. We consider the network from the end of training in Figure~\ref{fig:induction_heads} and perform the ablate-all-but-head-X ablation used for the circles in Figure~\ref{fig:additive_heads}. We then use our artificial optogenetics framework to ``perfect'' each head's attention pattern (so it would have an induction strength of 1). While this increases the accuracy for most heads, the accuracy is still far from perfect, indicating that inducion heads at the end of training are imperfect copiers. \textbf{b)} We dive deeper and find label dependence in the imperfect copying. Specifically, here we plot the relative mistake rate by output label for each head. Line opacity is equal to induction head strength, to emphasize the more prominent induction heads. Interestingly, Head 3 is relatively worse at copying labels 1 and 3, but heads 1 and 5 compensate for this. Notably, in Figure~\ref{fig:induction_heads}, heads 1 and 5 appear after Head 3. We postulate that heads 1 and 5 are thus compensating for the mistakes that Head 3 makes. Future work could further explore such a ``residual heads'' hypothesis in more depth. \textbf{c)} For completeness, we include accuracy vs output class. We note that Head 3 with a ``perfect'' attention pattern overall does make very few mistakes, as we'd expect given its high accuracy in a). Notably, what plot b shows is that Head 3 with a ``perfect'' attention pattern, when it does make mistakes, tends to do so on classes 1 and 3.
    }
    \label{fig:ih_imperfect_copier}
\end{figure}

Next, we present some evidence that supports the hypothesis. Namely, we consider perfecting each induction head's match circuit (by setting its attention pattern to the perfect pattern that gives attention 1 to the correct token and 0 to all other tokens), which allows us to quantify the quality of its copy circuit. Then, we can examine the mistake rate of the copy circuit for different labels, and see these values across heads. Our results are summarized in Figure~\ref{fig:ih_imperfect_copier}. Specifically, we do find some specialization of output classes: Head 3 has a higher mistake rate for copying labels 1 and 3, while heads 1 and 5 have virtually 0 mistake rate on these, indicating that they may have specialized to these examples. This result \textit{supports} the ``residual heads'' hypothesis.

If we repeat the analysis of Table \ref{tab:residual_head_test}, we find some promising evidence in this ``perfected match'' condition (Table \ref{tab:residual_head_test_perfect_copy}). These results suggest that, when it comes to the copying subcircuit, induction heads forming later in training preferentially focus on copying labels that earlier formed induction heads are worse at. As a result, these earlier heads do not have to ``fix their mistakes'' as much as if they were trained on their own, connecting to our results in Section~\ref{sec:networks_use_capacity}. This result is \textit{inconclusive} with respect to the ``residual heads'' hypothesis, as it only pertains to Head 1, and not the other heads.

\begin{figure}[h]
    \centering
    \includegraphics[width=0.7\textwidth]{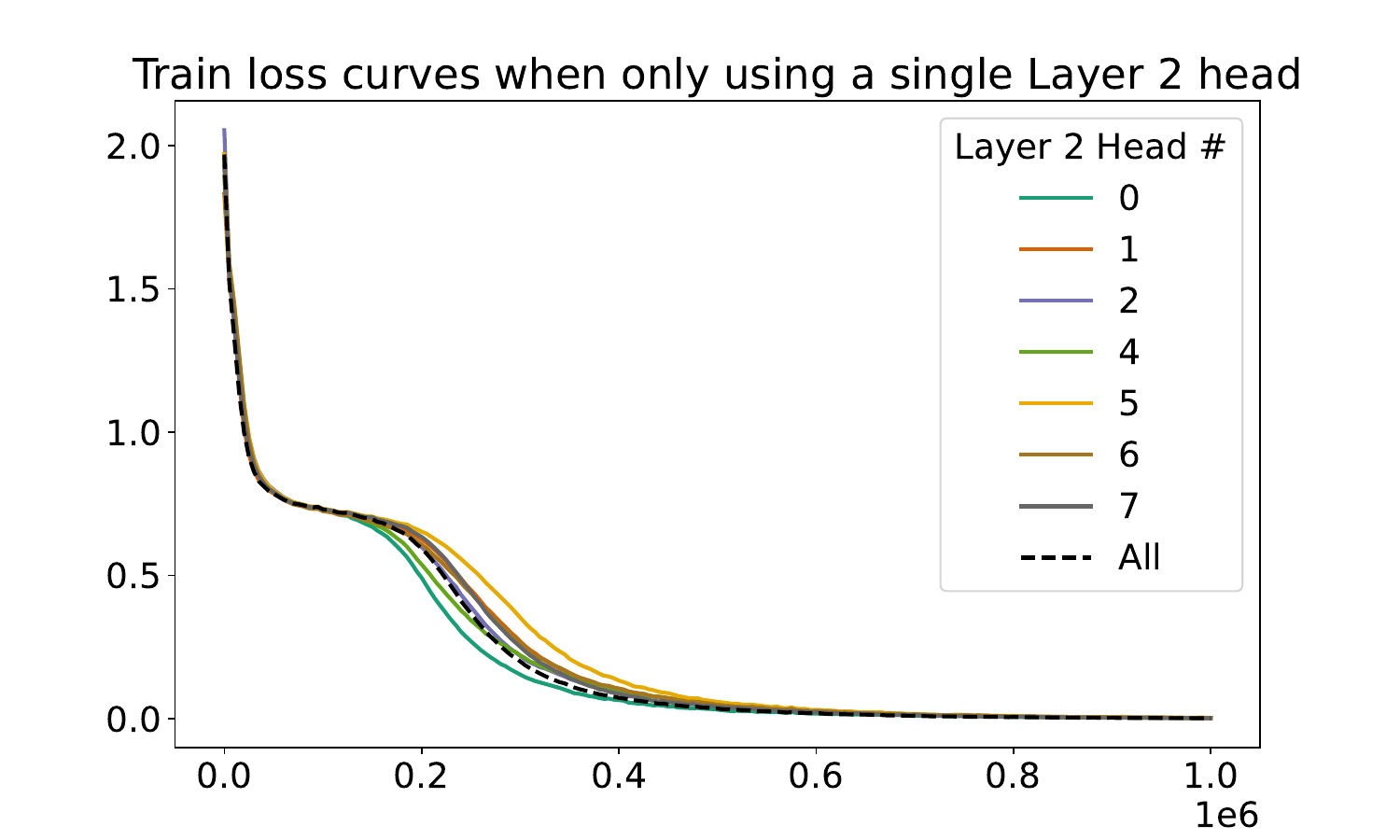}
    \caption{A follow-up to Figure~\ref{fig:additive_heads}b where we consider training with only a pair of Layer 2 heads active: Head 3 and one other head. Interestingly, loss curves snap to almost the true training loss curve. The relative ordering of phase changes is not indicative of the order of emergence (see Figure~\ref{fig:induction_heads}).}
    \label{fig:residual_head_withh3}
\end{figure}

Finally, we build off the result in Figure~\ref{fig:additive_heads}b, and try to take the argument further. Given that Head 3 is the quickest to learn and does in fact emerge first in the full network (Figure~\ref{fig:induction_heads}b), we consider training networks where all but \textit{two} heads are ablated: Head 3 and each of the other 7 heads (one at a time). Loss curves are shown in Figure~\ref{fig:residual_head_withh3}. By training with Head 3 and just one other head, we see loss curves are way closer to the actual training curve. However, the ordering of phase changes still does not match the ordering of emergence and strengths in Figure~\ref{fig:induction_heads}b, perhaps due to higher order interactions (e.g., between three heads or more). Given the current results though, this piece of evidence is \textit{against} the ``residual heads'' hypothesis.

Given all these pieces of evidence, for and against, we cannot convincingly argue for or against the ``residual heads'' hypothesis. We hope this investigation (including negative results) can inform future work researching this hypothesis.

\section{Reproducibility across model initializations}

Our code, as well as notebooks reproducing findings on multiple initialization seeds (5 seeds) can be found at \url{https://github.com/aadityasingh/icl-dynamics}. We found little to no differences when training seed (responsible for data ordering) was varied. Here, we summarize qualitatively the results that differ slightly across init seeds and the ones that don't (rather than providing $4 \cdot 6 = 30$ new figures).

\begin{figure}
    \centering
    \includegraphics[width=\textwidth]{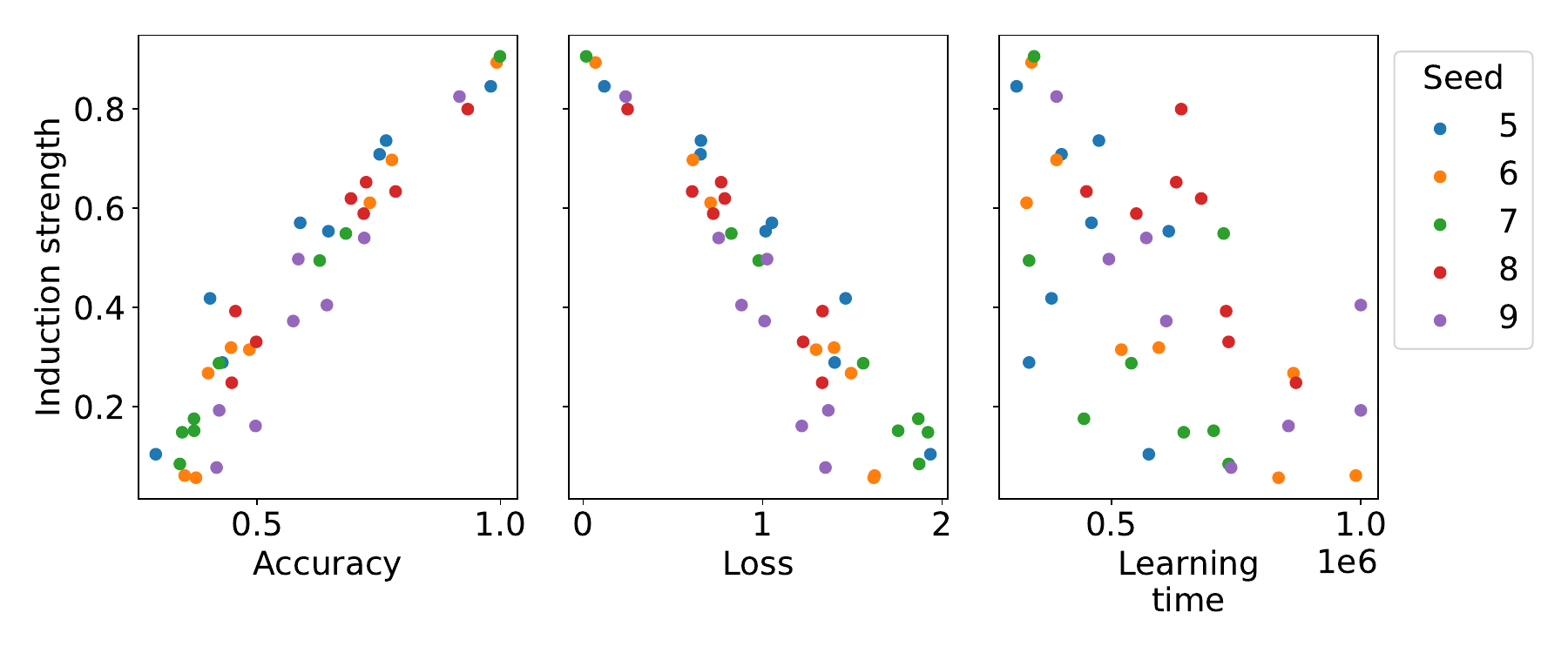}
    \caption{Reproducing some of the results of Section~\ref{sec:additive_heads} across seeds. Specifically, we plot the induction strength of all 8 L2 heads across 5 seeds (40 points) versus various other metrics on individual heads: accuracy and loss if all other L2 heads are ablated (analogous to circles in Figure~\ref{fig:additive_heads}a,\ref{fig:additive_heads_loss}), as well as the ``learning time'', which we define as the number of sequences seen such that the loss drops below $0.4\cdot \log 2$ when training with a single head (a summary statistic for Figure~\ref{fig:additive_heads}b). As we can see, induction head strength is well correlated with accuracy and loss (lower loss is better, and corresponds to stronger heads) across seeds. We see that learning time (and thus, connections to the neural race reduction \cite{saxe2022neural}) is correlated to strength across seeds (faster learning times correlate to stronger heads), but not as strongly. Future work could investigate these ideas of circuits ``racing'' to minimize the loss further.}
    \label{fig:seeds}
\end{figure}

Figure~\ref{fig:seeds} summarizes a reproduction of some of the results of Section~\ref{sec:additive_heads}. Specifically, strongest heads (determined via induction strength) are mostly able to solve the task. We find that the connection to the \textit{neural race} \cite{saxe2022neural} mostly holds across seeds (first discussed in Section~\ref{sec:networks_use_capacity}, though the correlation is not as strong as other metrics related to induction strength. Our main results on sub-circuit discovery (Figure~\ref{fig:subcircuits}) hold across seeds. The scaling result (Figure~\ref{fig:data_dependent}) reproduces across most seeds, though sometimes in standard training we see learning with 15 labels to be as quick as learning with 10 labels. For further details, we refer readers to the notebooks in our codebase.


\end{document}